%% file: rnns_10pages.tex
\documentclass[11pt]{article}
\usepackage{etex}
\usepackage{acl2015}
\usepackage{times}
\usepackage{url}
\usepackage{latexsym}

\usepackage{graphicx}
\usepackage{tikz}
\usepackage{tikz-qtree}
\usepackage{url}
\usepackage{caption}
\usepackage{todonotes}
\usepackage{pgfplots}
\usepackage{xcolor,colortbl}
\usepackage{stfloats}
\usepackage{amsmath}
 \usepackage{pstricks}
\usepackage{epstopdf}
\usepackage{multirow}

\usepackage{comment}

\usepackage[T1]{fontenc}
\usepackage{array}
\usepackage{makecell}

\usepackage{array}
\usepackage{makecell}
\newcolumntype{x}[1]{>{\centering\arraybackslash}p{#1}}

\newcommand\diag[4]{%
  \multicolumn{1}{p{#2}|}{\hskip-\tabcolsep
  $\vcenter{\begin{tikzpicture}[baseline=0,anchor=south west,inner sep=#1]
  \path[use as bounding box] (0,0) rectangle (#2+2\tabcolsep,\baselineskip);
  \node[minimum width={#2+2\tabcolsep},minimum height=\baselineskip+\extrarowheight] (box) {};
  \draw (box.north west) -- (box.south east);
  \node[anchor=south west] at (box.south west) {#3};
  \node[anchor=north east] at (box.north east) {#4};
 \end{tikzpicture}}$\hskip-\tabcolsep}}

\definecolor{Gray}{gray}{0.8}

\title{Toward Mention Detection Robustness with Recurrent Neural Networks}

\author{Thien Huu Nguyen$^\dag$\thanks{Work carried out during an internship at IBM},  Avirup Sil$^\S$, Georgiana Dinu$^\S$ and Radu Florian$^\S$ \\
 $^\dag$ Computer Science Department, New York University, New York, USA \\
 $^\S$ IBM T.J. Watson Research Center, Yorktown Heights, New York, USA\\
 {\tt 
thien@cs.nyu.edu,\{avi,gdinu,raduf\}.us.ibm.com}
 \\}
 
\date{}

\begin{document}
\maketitle

\begin{abstract}

One of the key challenges in natural language processing (NLP) is to yield good performance across application domains and languages. In this work, we investigate the robustness of the mention detection systems, one of the fundamental tasks in information extraction, via recurrent neural networks (RNNs). The advantage of RNNs over the traditional approaches is their capacity to capture long ranges of context and implicitly adapt the word embeddings, trained on a large corpus, into a task-specific word representation, but still preserve the original semantic generalization to be helpful across domains. Our systematic evaluation for RNN architectures demonstrates that RNNs not only outperform the best reported systems (up to 9\% relative error reduction) in the general setting but also achieve the state-of-the-art performance in the cross-domain setting for English. Regarding other languages, RNNs are significantly better than the traditional methods on the similar task of named entity recognition for Dutch (up to 22\% relative error reduction).


\end{abstract}

\section{Introduction}

One of the crucial steps toward understanding natural languages is mention detection (MD), whose goal is to identify entity mentions, whether named, nominal ({\it the president}) or pronominal ({\it he, she}), and classify them into some predefined types of interest in text such as PERSON, ORGANIZATION or LOCATION. This is an extension of the named entity recognition (NER) task which only aims to extract entity names. MD is necessary for many higher-level applications such as relation extraction, knowledge population, information retrieval, question answering and so on.

Traditionally, both MD and NER are formalized as sequential labeling problems, thereby being solved by some linear graphical models such as Hidden Markov Models (HMMs), Maximum Entropy Markov Models (MEMMs) or Conditional Random Fields (CRFs) \cite{Lafferty:01}. Although these graphical models have been adopted well to achieve the top performance for MD, there are still at least three problems we want to focus in this work:

(i) The first problem is the performance loss of the mention detectors when they are trained on some domain (the source domain) and applied to other domains (the target domains). The problem might originate from various mismatches between the source and the target domains (domain shifts) such as the vocabulary difference, the distribution mismatches etc \cite{Blitzer:06,Daume:07,Plank:13}.

(ii) Second, in mention detection, we might need to capture a long context, possibly covering the whole sentence, to correctly predict the type for a word. For instance, consider the following sentence with the pronominal ``{\it they}'':

{\it Now, the reason that \underline{France}, \underline{Russia} and \underline{Germany} are against war is because \underline{they} have suffered much from the past war.}

In this sentence, the correct type GPE\footnote{Geographical Political Entity} for ``{\it they}'' can only be inferred from its GPE references: ``{\it France}'', ``{\it Russia}'' and ``{\it Germany}'' which are far way from the pronominal ``{\it they}'' of interest. The challenge is come up with the models that can encode and utilize these long-range dependency context effectively.

(iii) The third challenge is to be able to quickly adapt the current techniques for MD so that they can perform well on new languages.

In this paper, we propose to address these problems for MD via recurrent neural networks (RNNs) which offer a decent recurrent mechanism to embed the sentence context into a distributed representation and employ it to decode the sentences. Besides, as RNNs replace the symbolic forms of words in the sentences with their word embeddings, the distributed representation that captures the general syntactic and semantic properties of words \cite{Collobert:08,Mnih:08,Turian:10}, they can alleviate the lexical sparsity, induce more general feature representation, thus generalizing well across domains \cite{Nguyen:15b}. This also helps RNNs to quickly and effectively adapt to new languages which just require word embeddings as the only new knowledge we need to obtain. Finally, we can achieve the task-specific word embeddings for MD to improve the overall performance by updating the initial pre-trained word embeddings during the course of training in RNNs.

The recent emerging interest in deep learning has produced many successful applications of RNNs for  NLP problems such as machine translation \cite{Cho:14a,Bahdanau:15}, semantic role labeling \cite{Zhou:15} etc. However, to the best of our knowledge, there has been no previous work employing RNNs for MD on the cross-domain and language settings so far. To summarize, the main contributions of this paper are as follow:

1. We perform a systematic investigation on various RNN architectures and word embedding techniques that are motivated from linguistic observations for MD.

2. We achieve the state-of-the-art performance for MD both in the general setting and in the cross-domain setting with the bidirectional modeling applied to RNNs.

3. We demonstrate the portability of the RNN models for MD to new languages by their significant improvement with large margins  over the best reported system for named entity recognition in Dutch.

\section{Related Work}

Both named entity recognition \cite{Bikel:97,Borthwick:97,Sang:03,Florian:03,Miller:04,Ando:05,Suzuki:08,Ratinov:09,Lin:09,Turian:10,Ritter:11,Passos:14,Cherry:15} and mention detection \cite{Florian:04} have been extensively studied with various evaluation in the last decades: MUC6, MUC7, CoNLL'02, CoNLL'03 and ACE. The previous work on MD has examined the cascade models \cite{Florian:06}, transferred knowledge from rich-resource languages to low-resource ones via machine translation \cite{Zitouni:08} or improved the systems on noisy input \cite{Florian:10}. Besides, some recent work also tries to solve MD jointly with other tasks such as relation or event extraction to benefit from their inter-dependencies \cite{Roth:07,Kate:10,Li:14a,Li:14b}. However, none of these work investigates RNNs for MD on the cross-domain and language settings as we do in this paper.


Regarding neural networks, a large volume of work has devoted to the application of deep learning to NLP in the last few years, centering around several network architecture such as convolutional neural networks (CNNs) \cite{Yih:14,Shen:14,Kalchbrenner:14,Kim:14,Zeng:14,Santos:15a,Santos:15b}, recurrent/recursive neural networks \cite{Socher:12,Cho:14a,Bahdanau:15,Zhou:15,Tai:15}, to name a few. For NER, Collobert et al. \shortcite{Collobert:11} propose a CNN-based framework while Mesnil et al. \shortcite{Mesnil:13} and Yao et al. \shortcite{Yao:13,Yao:14} investigate the RNNs for the slot filling problem in spoken language understanding. Although our work also examines the RNNs, we consider the mention detection problem with an emphasis on the robustness of the models in the domain shifts and language changes which has never been explored in the literature before.

Finally, for the robustness in the domain adaptation setting, the early work has focused on the sequential labeling tasks such as part-of-speech tagging or name tagging \cite{Blitzer:06,Huang:10,Daume:07,Xiao:13,Schnabel:14}. Recent work has drawn attention to relation extraction \cite{Plank:13,Nguyen:15a,Gormley:15}. In the field of neural networks, to the best of our knowledge, there is only one work from Nguyen and Grishman \shortcite{Nguyen:15b} that evaluates CNNs for event detection in the cross-domain setting.

\section{Models}

\label{sec:model}

We formalize the mention detection problem as a sequential labeling task. Given a sentence $X = w_1w_2\ldots w_n$, where $w_i$ is the $i$-th word and $n$ is the length of the sentence, we want to predict the label sequence $Y = y_1y_2\ldots y_n$ for $X$, where $y_i$ is the label for $w_i$. The labels $y_i$ follow the BIO2 encoding to capture the entity mentions in $X$ \cite{Ratinov:09}.

In order to prepare the sentence for RNNs, we first transform each word $w_i$ into a real-valued vector using the concatenation of two vectors $e_i$ and $f_i$: $w_i = [e_i, f_i]$\footnote{For simplicity, we are using the word $w_i$ and its real-valued vector representation interchangeably.}, where:

\begin{itemize}
\item $e_i$ is the word embedding vector of $w_i$, obtained by training a language model on a large corpus (discussed later).
\item $f_i$ is a binary vector encompassing different features for $w_i$. In this work, we are utilizing four types of features: capitalization, gazetteers, triggers (whether $w_i$ is present in a list of trigger words\footnote{Trigger words are the words that are often followed by entity names in sentences such as ``{\it president}'', ``{\it Mr.}'' etc.} or not) and cache (the label that is assigned to $w_i$ sometime before in the document).
\end{itemize}


We then enrich this vector representation by including the word vectors in a context window of $v_c$ for each word in the sentence to capture the short-term dependencies for prediction \cite{Mesnil:13}. This effectively converts $w_i$ into the context window version of the concatenated vectors: $x_i = [w_{i-v_c},\ldots,w_i,\ldots,w_{i+v_c}]$.

Given the new input representation, we describe the RNNs to be investigated in this work below.



\subsection{The Basic Models}

In standard recurrent neural networks, at each time step (word position in sentence) $i$, we have three main vectors: the input vector $x_i \in \mathbf{R}^I$, the hidden vector $h_i \in \mathbf{R}^H$ and the output vector $o_i \in \mathbf{R}^O$ ($I$, $H$ and $O$ are the dimensions of the input vectors, the dimension of the hidden vectors and the number of possible labels for each word respectively). The output vector $o_i$ is the probabilistic distribution over the possible labels for the word $x_i$ and obtained from $h_i$ via the softmax function $\varphi$:
$$
o_i = \varphi(W h_i),  \text{\hspace{1.2cm}} \varphi(z_m) = \frac{e^{z_m}}{\sum_k e^{z_k}}
$$

Regarding the hidden vectors or units $h_i$, there are two major methods to obtain them from the current input and the last hidden and output vectors, leading to two different RNN variants:

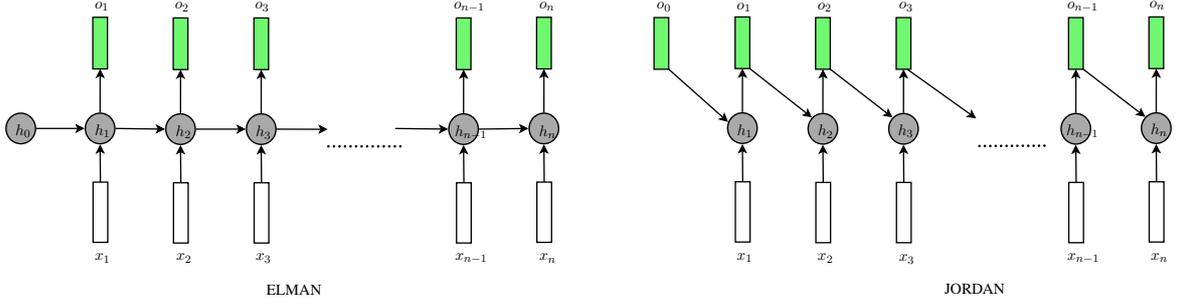
\begin{figure*}[!htb]
\centering
\input{elman_jordan.tex}
\caption{The ELMAN and JORDAN models}
\label{fig:basic}
\end{figure*}

\begin{itemize}
\item In the Elman model \cite{Elman:90}, called \textbf{ELMAN}, the hidden vector from the previous step $h_{i-1}$, along with the input in the current step $x_i$, constitute the inputs to compute the current hidden state $h_i$:
\begin{align}
\label{eq:1}
h_i = \Phi(Ux_i + Vh_{i-1})
\end{align}

\item In the Jordan model \cite{Jordan:86}, called \textbf{JORDAN}, the output vector from the previous step $o_{i-1}$ is fed into the current hidden layer rather than the hidden vector from the previous steps $h_{i-1}$. The rationale in this topology is to introduce the label from the preceding step as a feature for current prediction:
\begin{align}
\label{eq:2}
h_i = \Phi(Ux_i + Vo_{i-1})
\end{align}
\end{itemize}

In the formula above, $\Phi$ is the sigmoid activation function: $\Phi(z) = \frac{1}{1+e^{-z}}$ and $W$, $U$, $V$ are the same weight matrices for all time steps, to be learned during training. The unfolded dependency graphs for the two models are given in Figure \ref{fig:basic}.

\subsection{Gated Recurrent Units}

The hidden units in the two basic models above are essentially the standard feed-forward neural networks that take the vectors $h_{i-1}$, $o_{i-1}$ and $x_i$ as inputs and do a linear transformation followed by a nonlinearity to generate the hidden vector $h_i$. The ELMAN and JORDAN models are then basically a stack of these hidden units. Unfortunately, this staking mechanism causes the so-called ``{\it vanishing gradient}'' and ``{\it exploding gradient}'' problems \cite{Bengio:94}, making it challenging to train the networks properly in practice \cite{Pascanu:12}. These problems are addressed by the long-short term memory units (LSTM) \cite{Hochreiter:97,Graves:09} that propose the idea of memory cells with four gates to allow the information storage and access over a long period of time. 

In this work, we apply another version of memory units with only two gates ({\it reset} and {\it update}), called {\it Gated Recurrent Units} (GRUs) from Cho et al. \shortcite{Cho:14a,Bahdanau:15}. GRU is shown to be much simpler than LSTM in terms of computation and implementation but still achieves the comparable performance in machine translation \cite{Bahdanau:15}.

The introduction of GRUs into the models ELMAN and JORDAN amounts to two new models, named \textbf{ELMAN\_GRU} and \textbf{JORDAN\_GRU} respectively, with two new methods to compute the hidden vectors $h_i$. The formula for ELMAN\_GRU is adopted directly from Cho et al. \shortcite{Cho:14b} and given below:

$$
h_i = z_i \odot \hat{h}_i + (1 - z_i) \odot h_{i-1}
$$
$$
\hat{h}_i = \Phi(W_h x_i + U_h(r_i \odot h_{i-1}))
$$
\begin{align}
\label{eq:3}
z_i = \Phi(W_zx_i + U_z h_{i-1})
\end{align}
$$
r_i = \Phi(W_r x_i + U_r h_{i-1})
$$

where $W_h, W_z, W_r \in \mathbf{R}^{H \times I}$, $U_h, U_z, U_r \in \mathbf{R}^{H \times H}$ and $\odot$ is the element-wise multiplication operation.


We cannot directly apply the formula above to the JORDAN\_GRU model since the dimensions of the output vectors $o_i$ and the hidden vector $h_i$ are different in general. For JORDAN\_GRU, we first need to transform the output vector $o_i$ into the hidden vector space, leading to the following formula:

$$
h_i = z_i \odot \hat{o}_i + (1 - z_i) \odot t_{i-1}
$$
$$
t_{i-1} = To_{i-1}
$$
$$
\hat{o}_i = \Phi(W_o x_i + U_o(r_i \odot t_{i-1}))
$$
\begin{align}
\label{eq:4}
z_i = \Phi(W_zx_i + U_z t_{i-1})
\end{align}
$$
r_i = \Phi(W_r x_i + U_r t_{i-1})
$$

where $T \in \mathbf{R}^{H \times O}$.

\subsection{The Extended Networks}

One of the limitations of the four basic models presented above is their incapacity to incorporate the future context information that might be crucial to the prediction in the current step. For instance, consider the first word ``{\it Liverpool}'' in the following sentence:

{\it \underline{\textbf{Liverpool}} suffered an upset first home league defeat of the season, beaten 1-0 by a Guy Whittingham goal for Sheffield Wednesday.}

In this case, the correct label ORGANIZATION can only be detected if we first go over the whole sentence and then utilize the context words after ``{\it Liverpool}'' to decide its label.

The limitation of the four models is originated from their mechanism to perform a single pass over the sentences from left to right and make the prediction for a word once they first encounter it. In the following, we investigate two different networks to overcome this limitation.

\subsubsection{The Contextual Networks}

\begin{figure*}[!htb]
\centering
\input{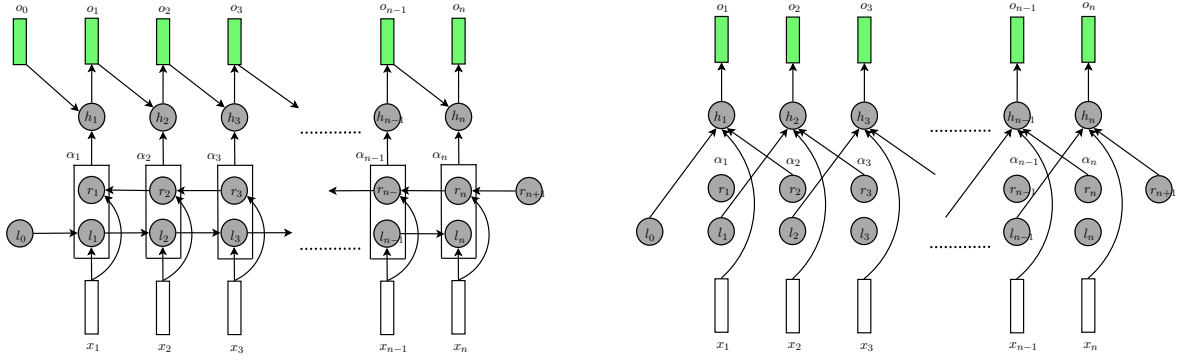}
\caption{The bidirectional models. The model on the right is from Mesnil et al. \shortcite{Mesnil:13} with the forward and backward context size of 1.  $l_0, r_{n+1}$ are the zero vectors.}
\label{fig:bidirection}
\end{figure*}

The contextual networks are motivated by the RNN Encoder-Decoder models that have become very popular in neural machine translation recently \cite{Cho:14a,Bahdanau:15}. In these networks, we first run a RNN $R_e$ over the whole sentence $X = x_1x_2\ldots x_n$ to collect the hidden vector sequence $c_1, c_2, \ldots, c_n$, where $c_i$ is the hidden vector for the $i$-th step in the sentence. For convenience, this process is denoted by:

$$
R_e(x_1x_2\ldots x_n) = c_1,c_2,\ldots, c_n
$$

The final hidden vector $c_n$ is then considered as a distributed representation of $X$, encoding the global context or topic information for $X$ (the encoding phrase) and thus possibly being helpful for the label prediction of each word in $X$. Consequently, we perform the second RNN $R_d$ over $X$ to decode the sentence in which $c_n$ is used as an additional input in computing the hidden units for every time step (the decoding phrase).

Note that $R_e$ (the encoding model) should be an Elman model\footnote{From now on, for convenience, the term ``{\it Elman models}'' refers to the ELMAN and ELMAN\_GRU models. The same implication applies for the Jordan models.} while $R_d$ (the decoding model) can be any Elman or Jordan model. As an example, the formula for $R_d$ = ELMAN is:



$$
h_i = \Phi(Ux_i + Vh_{i-1} + Sc_n)
$$



\subsubsection{The Bidirectional Networks}
\label{subsec:bidirect}

The bidirectional networks involve three passes over the sentence, in which the   first two passes are designated to encode the sentence while the third pass is responsible for decoding. The procedure for the sentence $X = x_1x_2\ldots x_n$ is below: \\

(i) Run the first RNN $R_{ef}$ from left to right over $x_1x_2\ldots x_n$ to obtain the first hidden vector or output vector sequence (depending on whether $R_{ef}$ is an Elman or Jordan network respectively): $R_{ef}(x_1x_2\ldots x_n) = l_1,l_2,\ldots, l_n$ (forward encoding). 

(ii) Run the second RNN $R_{eb}$ from right to left over $x_1x_2\ldots x_n$ to obtain the second hidden vector or output vector sequence: $R_{eb}(x_nx_{n-1}\ldots x_1) = r_n,r_{n-1},\ldots, r_1$ (backward encoding).

(iii) Obtain the concatenated sequence $\alpha = \alpha_1,\alpha_2,\ldots, \alpha_n$ where $\alpha_i = [l_i, r_i]$.

(iv) Decode the sentence with the third RNN $R_d$ (the decoding model) using $\alpha$ as the input vector, i.e, replacing $x_i$ by $\alpha_i$ in the formula (\ref{eq:1}), (\ref{eq:2}), (\ref{eq:3}) and (\ref{eq:4}). \\

Conceptually, the encoding RNNs $R_{ef}$ and $R_{eb}$ can be different but in this work, for simplicity and consistency, we assume that we only have a single encoding model, i.e, $R_{ef} = R_{eb} = R_e$. Once again, $R_e$ and $R_d$ can be any model in \{ELMAN, JORDAN, ELMAN\_GRU, JORDAN\_GRU\}.


The observation is, at the time step $i$, the forward hidden vector $l_i$ represents the encoding for the past word context (from position 1 to $i$) while the backward hidden vector $r_i$ is the summary for the future word context (from position $n$ to $i$). Consequently, the concatenated vector $\alpha_i = [l_i, r_i]$ constitutes a distributed representation that is specific to the word at position $i$ but still encapsulates the context information over the whole sentence at the same time. This effectively provides the networks a much richer representation to decode the sentence. The bidirectional network for $R_e$ = ELMAN and $R_d$ = JORDAN is given on the left of Figure \ref{fig:bidirection}.

We notice that Mesnil et al. \shortcite{Mesnil:13} also investigate the bidirectional models for the task of slot filling in spoken language understanding. However, compared to the work presented here, Mesnil et al. \shortcite{Mesnil:13} does not use any special transition memory cells (like the GRUs we are employing in this paper) to avoid numerical stability issues \cite{Pascanu:12}. Besides, they form the inputs $\alpha$ for the decoding phase from a larger context of the forward and backward encoding outputs, while performing word-wise, independent classification; in contrast, we use only the current output vectors in the forward and backward encodings for $\alpha$, but perform recursive computations to decode the sentence via the RNN model $R_d$ (demonstrated on the right of Figure \ref{fig:bidirection}).

\subsection{Training and Inference}

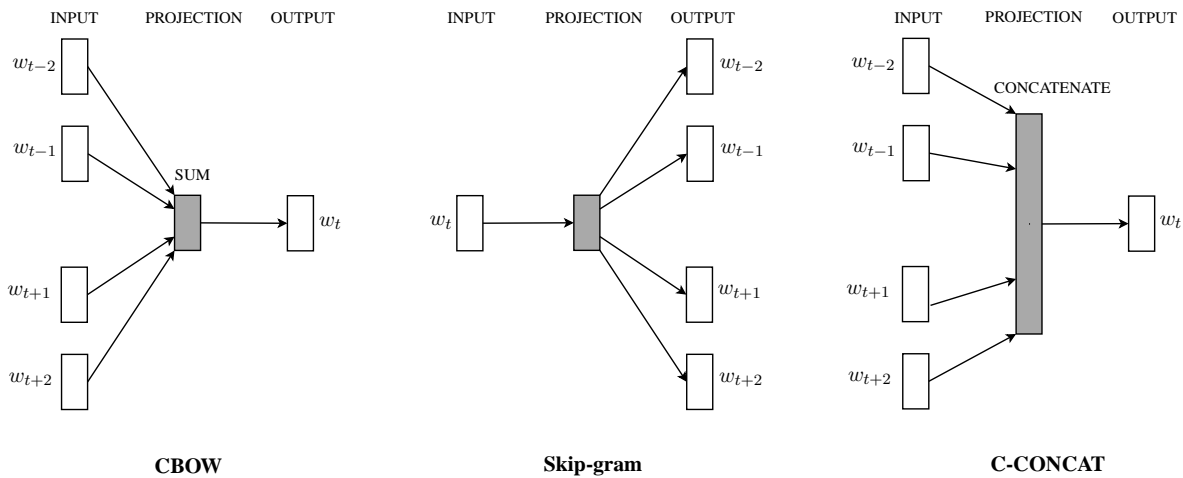
\begin{figure*}[!htb]
\centering
{\input{wordEmbeddings.tex}}
\caption{Methods to Train Word Embeddings}
\label{fig:wordEmbeddings}
\end{figure*}

We train the networks locally. In particular, each training example consists of a word $x_i$ and its corresponding label $y_i$ in a sentence $X = x_1x_2\ldots x_n$ (denoted by $E = (x_i, y_i, X)$). In the encoding phase, we first compute the necessary inputs according to the specific model of interest. This can be the original input vectors $x_1,x_2,\ldots, x_n$ in the four basic models or the concatenated vectors $\alpha_1,\alpha_2,\ldots, \alpha_n$ in the bidirectional models. For the contextual models, we additionally have the context vector $c_n$. Eventually, in the decoding phase, an sequence of $v_d$ input vectors preceding the current position $i$ is fed into the decoding network $R_d$ to obtain the output vector sequence. The last vector in this output sequence corresponds to the probabilistic label distribution for the current position $i$, to be used to compute the objective function. For example, in the bidirectional models, the input sequence for the decoding phase is $\alpha_{i-v_d}\alpha_{i-v_d+1}\ldots \alpha_i$ while the output sequence is: $
R_e({\alpha_{i-v_d}\alpha_{i-v_d+1}\ldots \alpha_i}) = o_{i-v_d}o_{i-v_d+1}\ldots o_i$.


In this work, we employ the stochastic gradient descent algorithm\footnote{We try the $AdaDelta$ algorithm \protected\cite{Zeiler:12} and the dropout regularization but do not see much difference.} to update the parameters via minimizing the negative log-likelihood objective function: $ \mathbf{nll}(E) = - \log(o_i[y_i]) $.

Finally, besides the weight matrices in the networks, the word embeddings are also optimized during training to obtain the task-specific word embeddings for MD. The gradients are computed using the back-propagation through time algorithm \cite{Mozer:89} and inference is performed by running the networks over the whole sentences and taking $\mathrm{argmax}$ over the output sequence: $y_i = \mathrm{argmax} (o_i)$.

\section{Word Representation}
\label{sec:wordRep}

Following Collobert et al. \shortcite{Collobert:11}, we pre-train word embeddings from a large corpus and employ them to initialize the word representations in the models. One of the state-of-the-art models to train word embeddings have been proposed recently in Mikolov et al. \shortcite{Mikolov:13a,Mikolov:13b} that introduce two log-linear models, i.e the continuous bag-of-words model (CBOW) and the continuous skip-gram model (Skip-gram). The CBOW model attempts to predict the current word based on {\it the average of the context word vectors} while the Skip-gram model aims to predict the surrounding words in a sentence given the current word. In this work, besides the CBOW and skip-gram models, we examine a concatenation-based variant of CBOW (C-CONCAT) to train word embeddings and compare the three models to gain insights into which kind of model is effective to obtain word representations for the MD task. The objective of C-CONCAT is to predict the target word using {\it the concatenation of the vectors of the words surrounding it}, motivated from our strategy to decide the label for a word based on the concatenated context vectors. Intuitively, the C-CONCAT model would perform better than CBOW due to the close relatedness between the decoding strategies of C-CONCAT and the MD methods. CBOW, Skip-gram and C-CONCAT are illustrated in Figure \ref{fig:wordEmbeddings}.


\section{Experiments}

\subsection{Dataset}

Our main focus in this work is to evaluate the robustness of the MD systems across domains and languages. In order to investigate the robustness across domains, following the prior work \cite{Plank:13,Nguyen:15a}, we utilize the ACE 2005 dataset which contains 6 domains: broadcast news ({\it bn}), newswire ({\it nw}), broadcast  conversation ({\it bc}), telephone conversation ({\it cts}), weblogs ({\it wl}), usenet ({\it un}) and 7 entity types: person, organization, GPE, location, facility, weapon, vehicle. The union of {\it bn} and {\it nw} is considered as a single domain, called \textbf{news}. We take half of {\it bc} as the only development data and use the remaining data and domains for evaluation. Some statistics about the domains are given in Table \ref{tab:ace}. As shown in Plank and Moschitti \shortcite{Plank:13}, the vocabulary of the domains is quite different.

For completeness, we also test the RNNs system on the Named Entity Recognition for English using the CoNLL 2003 dataset\footnote{\tiny \url{http://www.cnts.ua.ac.be/conll2003/ner/}} \cite{Florian:03,Sang:03} and compare the performance with the state-of-the-art neural network system on this dataset \cite{Collobert:11}. Regarding the robustness across languages, we further evaluate the RNN models on the CoNLL 2002 dataset for Dutch Named Entity Recognition\footnote{\tiny \url{http://www.cnts.ua.ac.be/conll2002/ner/}} \cite{Carreras:02,Sang:02}. Both CoNLL datasets come along with the training data, validation data and test data, annotated for 4 types of entities: person, organization, location and miscellaneous.

\begin{table}[htbp]
\small
\centering
\begin{tabular}{|lccc|}
\hline
Domain & \multicolumn{1}{c}{\#Docs} & \multicolumn{1}{c}{\#Sents} & \multicolumn{1}{c|}{\#Mentions} \\ \hline
news & 332 & 6487 & 22460 \\ 
bc & 60 & 3720 & 9336 \\ 
cts & 39 & 5900 & 9924 \\ 
wl & 119 & 2447 & 6538 \\ 
un & 49 & 2746 & 6507 \\ \hline
Total & 599 & 21300 & 54765 \\ \hline
\end{tabular}
\caption{ACE 2005 Dataset}
\label{tab:ace}
\end{table}


Finally, we use the standard IOB2 tagging schema for the ACE 2005 dataset and the Dutch CoNLL dataset while the IOBES tagging schema is applied for the English CoNLL dataset to ensure the compatibility with Collobert et al. \shortcite{Collobert:11}.

\subsection{Resources and Parameters}

In all the experiments for RNNs below, we employ the context window $v_c = 5$, the decoding window $v_d = 9$. We find that the optimal number of hidden units (or the dimension of the hidden vectors) and the learning rate vary according to the dataset. For the ACE 2005 dataset, we utilize 200 hidden units with learning rate = 0.01 while these numbers are 100 and 0.06 respectively for the CoNLL datasets. Note that the number of hidden units is kept the same in both the encoding phase and the decoding phase.

For word representation, we train the word embeddings for English from the Gigaword corpus augmented with the newsgroups data from BOLT (Broad Operational Language Technologies) (6 billion tokens) while the entire Dutch Wikipedia pages (310 million tokens) are extracted to train the Dutch word embeddings. We utilize the word2vec toolkit\footnote{\tiny \url{https://code.google.com/p/word2vec/}} (modified to add the C-CONCAT model) to learn the word representations. Following Baroni et al. \shortcite{Baroni:14}, we use the context window of 5, subsampling set to 1$e$-05 and negative sampling with the number of instances set to 10. The dimension of the vectors is set to 300 to make it comparable with the word2vec toolkit.

\subsection{Model Architecture Experiments}

\subsubsection{Model Architecture Evaluation}

In this section, we evaluate different RNN models by training the models on the \textbf{news} domain and report the performance on the development set. As presented in the previous sections, we have 4 basic models $M$ = \{ELMAN, JORDAN, ELMAN\_GRU, JORDAN\_GRU\}, 8 contextual models (two choices for the encoding model $R_e$ in \{ELMAN, ELMAN\_GRU\} and 4 choices for the decoding model $R_d \in M$), and 16 bidirectional models (4 choices for the encoding and decoding models $R_e$, $R_d$ in M). The performance for the basic models, the contextual models and the bidirectional models are shown in Table \ref{tab:basicDev}, Table \ref{tab:contextDev} and Table \ref{tab:bidirectionalDev} respectively\footnote{The experiments in this section use C-CONCAT to pre-train word embeddings.}.

\begin{table}[htbp]
\small
\centering
\begin{tabular}{lc}
Model($R_d$) & F1 \\ \hline
ELMAN & 80.70 \\ 
JORDAN & 80.46 \\ 
ELMAN\_GRU & 80.85 \\ 
JORDAN\_GRU & \textbf{81.06} \\ 
\end{tabular}
\caption{The basic models' performance}
\label{tab:basicDev}
\end{table}

\begin{table}[htbp]
\small
\centering
\begin{tabular}{l|cc}
\diag{.1em}{2.5cm}{$R_d$}{$R_e$} & ELMAN & ELMAN\_GRU \\ \hline
ELMAN & 80.38 & 80.16 \\ 
JORDAN & 80.67 & 80.66 \\ 
ELMAN\_GRU & 80.56 & 79.61 \\ 
JORDAN\_GRU & \textbf{80.77} & 79.77 \\ 
\end{tabular}
\caption{The contextual models' performance}
\label{tab:contextDev}
\end{table}

\begin{table}[htbp]
\small
\centering
\begin{tabular}{|l|cc}
\diag{.1em}{2.5cm}{$R_d$}{$R_e$} & ELMAN & ELMAN\_GRU \\ \hline
ELMAN & 80.99 & 81.42 \\ 
JORDAN & 81.14 & 81.68 \\ 
ELMAN\_GRU & 80.53 & 81.16 \\ 
JORDAN\_GRU & 80.98 & \textbf{82.37} \\ \hline
\diag{.1em}{2.5cm}{$R_d$}{$R_e$} & JORDAN & JORDAN\_GRU \\ \hline
ELMAN & 79.12 & 79.64 \\ 
JORDAN & 79.21 & 80.85 \\ 
ELMAN\_GRU & 79.80 & 80.41 \\ 
JORDAN\_GRU & 79.76 & 81.02 \\ 
\end{tabular}
\caption{The bidirectional models' performance}
\label{tab:bidirectionalDev}
\end{table}

There are several important observations from the three tables:

-Elman vs Jordan: In the encoding phase, the Elman models consistently outperform the Jordan models when the same decoding model is applied in the bidirectional architecture. In the decoding phase, however, it turns out that the Jordan models are better most of the time over different model architectures (basic, contextual or bidirectional). 


-With vs Without GRUs: The trends are quite mixed in the comparison between the cases with and without GRUs. In particular, for the encoding part, given the same decoding model, GRUs are very helpful in the bidirectional architecture while this is not the case for the contextual architecture. For the decoding part, we can only see the clear benefit of GRUs in the basic models and the bidirectional architecture when $R_e$ is a Jordan model.

-Regarding different model architectures, in general, the bidirectional models are more effective than the contextual models and the basic models, confirming the effectiveness of bidirectional modeling to achieve a richer representation for MD. 



The best basic model (F1 = 81.06\%), the best contextual model (F1 = 80.77\%) and the best bidirectional model with (F1 = 82.37\%) are called BASIC, CONTEXT and BIDIRECT respectively. In the following, we only focus on these best models in the experiments.

\subsubsection{Comparison to other Bidirectional RNN Work}

Mesnil et al. \shortcite{Mesnil:13} also present a RNN system with bidirectional modeling for the slot filling task. As described in Section \ref{subsec:bidirect}, the major difference between the bidirectional models in this work and Mesnil el al. \shortcite{Mesnil:13}'s is the recurrence in our decoding phase. Table \ref{tab:mesnil} compares the performance of the bidirectional model from Mesnil et al. \shortcite{Mesnil:13}, called MESNIL, and the  BIDIRECT model. In order to verify the effectiveness of recurrence in decoding, the performance of MESNIL incorporated with the JORDAN\_GRU model in the decoding phase (MESNIL+JORDAN\_GRU) is also reported.


\begin{table}[htbp]
\small
\centering
\begin{tabular}{l|c|c|c}
Model & P & R & F1 \\ \hline
\small MESNIL \shortcite{Mesnil:13}  & 81.01 & 79.67 & 80.33 \\
\small MESNIL + JORDAN\_GRU  & 82.17 & 79.56 & 80.85 \\ \hline
\small BIDIRECT & 82.91 & 81.83 & \textbf{82.37} \\
\end{tabular}
\caption{Comparison to Mesnil et al. \shortcite{Mesnil:13}.}
\label{tab:mesnil}
\end{table}

In general, we see that the bidirectional model in this work is much better than the model in Mesnil et al. \shortcite{Mesnil:13} for MD. This is significant with $p < 0.05$ and a large margin (an absolute improvement of 2.04\%). More interestingly, MESNIL is further improved when it is augmented with the JORDAN\_GRU decoding, verifying the importance of recurrence in decoding for MD. 

\subsection{Word Embedding Evaluation}

The section investigates the effectiveness of different techniques to learn word embeddings to initialize the RNNs for MD. Table \ref{tab:embedding} presents the performance of the BASIC, CONTEXT and BIDIRECT models on the development set (trained on news)  when the CBOW, SKIP-GRAM and C-CONCAT techniques are utilized to obtain word embeddings from the same English corpus. We also report the performance of the models when they are initialized with the word2vec word embeddings from Mikolov et al. \shortcite{Mikolov:13a,Mikolov:13b} (trained with the Skip-gram model on 100 billion words of Google News) (WORD2VEC). All of these word embeddings are updated during the training of the RNNs to induce the task-specific word embeddings . Finally, for comparison purpose, the performance for the two following scenarios is also included: (i) the word vectors are initialized randomly (not using any pre-trained word embeddings) (RANDOM), and (ii) the word vectors are loaded from the C-CONCAT pre-trained word embeddings but fixed during the RNN training (FIXED).

\begin{table}[htbp]
\centering
\small
\begin{tabular}{|l|c|c|c|}
\hline
Word & \multicolumn{ 3}{c|}{Model} \\ \cline{2-4}
Embeddings & \small BASIC & \small CONTEXT & \small BIDIRECT \\ \hline
\small RANDOM & 79.30 & 79.49 & 79.76 \\ \hline
\small FIXED & 80.36 & 80.60 & 81.52 \\ \hline
\small WORD2VEC & 80.92 & 81.66 & 81.41 \\ \hline
\small CBOW & 78.61 & 79.58 & 79.74 \\ \hline
\small SKIP-GRAM & 81.45 & 81.59 & 81.96 \\ \hline
\small C-CONCAT & 81.06 & 80.77 & \textbf{82.37} \\ \hline
\end{tabular}
\caption{Word Embedding Comparison}
\label{tab:embedding}
\end{table}

The first observation is that we need to borrow some pre-trained word embeddings and update them during the training process to improve the MD performance (comparing C-CONCAT, RANDOM and FIXED). Second, C-CONCAT is much better than CBOW, confirming our hypothesis about the similarity between the decodings of C-BOW and MD in Section \ref{sec:wordRep}. Third, we do not see much difference in terms of MD performance when we enlarge the corpus to learn word embeddings (comparing SKIP-GRAM and WORD2VEC that is trained with the skip-gram model on a much larger corpus). Finally, we achieve the best performance when we apply the C-CONCAT technique in the BIDIRECT model. From now on, for consistency, we use the C-CONCAT word embeddings in all the experiments below.

\subsection{Comparison to the State-of-the-art}

\subsubsection{The ACE 2005 Dataset for Mention Detection}

The state-of-the-art systems for mention detection on the ACE 2005 dataset have been the joint extraction system for entity mentions and relations from Li and Ji \shortcite{Li:14a} and the information networks to unify the outputs of three information extraction tasks: entity mentions, relations and events using structured perceptron from Li et al. \shortcite{Li:14b}. They extensively hand-design a large set of features (parsers, gazetteers, word clusters, coreference etc) to capture the inter-dependencies between different tasks. In this section, besides comparing the RNN systems above with these state-of-the-art systems, we also implement a Maximum Entropy Markov Model (MEMM) system\footnote{We tried the CRF model, but it is worse than MEMM in our case.}, following the description and features in Florian et al. \shortcite{Florian:04,Florian:06}, and include it in the comparison for completeness\footnote{We notice that the four features we are using in the RNN models (Section \ref{sec:model}) are also included in the feature set of the implemented MEMM system.}. For this comparison, following Li and Ji \shortcite{Li:14a}, we remove the documents from the two informal domains cts and un, and then randomly split the remaining 511 documents into 3 parts: 351 for training, 80 for development, and the rest 80 for blind test. The performance of the systems on the blind test set is presented in Table \ref{tab:acePerformance}.

\begin{table}[htbp]
\small
\centering
\begin{tabular}{l|c|c|c}
System & P & R & F1 \\ \hline
Joint System \cite{Li:14a}  & 85.2 & 76.9 & 80.8 \\
Joint System \cite{Li:14b}  & 85.1 & 77.3 & 81.0 \\ \hline
MEMM & 84.4 & 80.5 & 82.4 \\ \hline
BASIC & 83.4 & 80.8 & 82.1 \\
CONTEXT & 81.8 & 81.6 & 81.7 \\
BIDIRECT & 83.7 & 81.8 & \textbf{82.7}\\
\end{tabular}
\caption{Performance for MD on ACE.}
\label{tab:acePerformance}
\end{table}

There are two main conclusions from the table. First, our MEMM system is already better than the state-of-the-art system on this dataset, possibly due to the superiority of the features we are employing in this system. Consequently, from now on, we would utilize this MEMM system as the baseline in the following experiments. Second, all the three RNN systems: BASIC, CONTEXT and BIDIRECT substantially outperform the state-of-the-art system with large margins. In fact, we achieve the best performance on this dataset with the BIDIRECT model, once again testifying to the benefit of bidirectional modeling for MD.

\subsubsection{The CoNLL 2003 Dataset for English Named Entity Recognition}

This section further assess the RNN systems on the similar task of Named Entity Recognition for English to compare them with other neural network approaches for completeness. On the CoNLL 2003 dataset for English NER, the best neural network system so far is Collobert et al. \shortcite{Collobert:11}. This system, called CNN-Sentence, employs convolutional neural networks to encode the sentences and then decodes it at the sentence level. Table \ref{tab:englishCoNLL} shows the performance of CNN-Sentence and our RNN systems on this dataset.

\begin{table}[htbp]
\small
\centering
\begin{tabular}{l|c}
System & F1 \\ \hline
CNN-Sentence & 89.59 \\  \hline
BASIC & 89.26 \\ 
CONTEXT & 88.88 \\ 
BIDIRECT & \textbf{89.86} \\ 
\end{tabular}
\caption{Performance on English CoNLL 2003.}
\label{tab:englishCoNLL}
\end{table}

\begin{table*}[htbp]
\small
\centering
\resizebox{\textwidth}{!}{
\begin{tabular}{|l|c|c|c|c|c|c|c|c|c|c|}
\hline
System & \multicolumn{ 5}{c|}{Without Features} & \multicolumn{ 5}{c|}{With Features} \\ \cline{2-11}
 & In-Domain & bc & cts & wl & un & In-Domain & bc & cts & wl & un \\ \hline
MEMM & 76.90 & 71.73 & 78.02 & 66.89 & 67.77 & \textbf{82.55} & 78.33 & 87.17 & \textbf{76.70} & 76.75 \\ \hline
BASIC & 79.01 & \textbf{77.06} & 85.42 & 73.00 & 72.93 & 81.99 & 78.75 & 86.51 & 76.60 & 76.94 \\ \hline
CONTEXT & 78.27 & 73.55 & 84.85 & 73.39 & 72.26 & 81.61 & 77.84 & 87.79 & 76.60 & 76.41 \\ \hline
BIDIRECT & \textbf{80.00\dag} & 76.27\dag & \textbf{85.64\dag} & \textbf{73.79\dag} & \textbf{73.88\dag} & 82.52 & \textbf{79.65\dag} & \textbf{88.43\dag} & \textbf{76.70} & \textbf{77.03} \\ \hline
\end{tabular}
}
\caption{System's Performance on the Cross-domain Setting. Cells marked with \dag  designate the BIDIRECT models that significantly outperform ($p < 0.05$) the MEMM model on the specified domains.}
\label{tab:crossNews}
\end{table*}

\begin{table*}[htbp]
\small
\centering
\resizebox{\textwidth}{!}{
\begin{tabular}{|l|c|c|c|c|c|c|c|c|c|c|c|c|}
\hline
& \multicolumn{4}{c|}{MEMM} & \multicolumn{4}{c|}{BIDIRECT} & \multicolumn{4}{c|}{BIDIRECT-MEMM} \\ \cline{2-13}
 & bc & cts & wl & un & bc & cts & wl & un & bc & cts & wl & un \\ \hline
bc & 75.20 & 86.60 & 70.25 & 72.38 & 75.49 & 87.51 & 70.75 & 73.04 & \textbf{0.29} & \textbf{0.91\dag} & \textbf{0.50\dag} & \textbf{0.66\dag} \\ \hline
cts & 66.91 & 89.76 & 68.74 & 69.72 & 68.23 & 91.24 & 68.82 & 70.27 & \textbf{1.32\dag} & \textbf{1.48\dag} & \textbf{0.08} & \textbf{0.55\dag} \\ \hline
wl & 74.94 & 86.53 & 77.07 & 75.90 & 74.73 & 86.79 & 76.35 & 75.37 & -0.21 & \textbf{0.26} & -0.72 & -0.53 \\ \hline
un & 72.72 & 86.75 & 72.04 & 73.47 & 73.53 & 88.29 & 73.16 & 74.00 & \textbf{0.81\dag} & \textbf{1.45\dag} & \textbf{1.12\dag} & \textbf{0.53\dag} \\ \hline
\end{tabular}
}
\caption{Comparison between MEMM and BIDIRECT. Cells marked with \dag  designate the statistical significance ($p < 0.05$). The columns and rows correspond to the source and target domains respectively.}
\label{tab:crossAll}
\end{table*}

As we can see from the table, the RNN systems are on par with the CNN-Sentence system from Collobert et al. \shortcite{Collobert:11} except the CONTEXT system that is worse in this case. We actually accomplish the best performance with the BIDIRECT model, thus further demonstrating its virtue.

\subsection{Cross-Domain Experiments}

One of the main problems we want to address in this work is the robustness across domains of the MD systems. This section tests the MEMM (the baseline) and the RNN systems on the cross-domain settings to gain an insights into their operation when the domain changes. In particular, for the first experiment, following the previous work of domain adaptation on the ACE 2005 dataset \cite{Plank:13,Nguyen:14,Nguyen:15a}, we treat {\it news} as the source domain and the other domains: {\it bc}, {\it cts}, {\it wl} and {\it un} as the target domains. We then examine the systems on two scenarios: (i) the systems are trained and tested on the source domain via 5-fold cross validation (in-domain performance), and (ii) the systems are trained on the source domain but evaluated on the target domains. Besides, in order to understand the effect of the features on the systems, we report the systems' performance in both the inclusion and exclusion of the features described in Section \ref{sec:model}. Table \ref{tab:crossNews} presents the results.

To summarize, we find that the RNN systems significantly outperform the MEMM system across all the target domains when the features are not applied. The BIDIRECT system still yields the best performance among systems being investigated (except in domain {\it bc}). This is also the case in the inclusion of features and demonstrates the robustness of the BIDIRECT model in the domain shifts. We further support this result in Table \ref{tab:crossAll} where we report the performance of the MEMM and BIDIRECT systems (with features) on different domain assignments for the source and the target domains. Finally, we also see that the features are very useful for both the MEMM and the RNNs.

\subsection{Named Entity Recognition for Dutch}

The previous sections have dealt with mention detection for English. In this section, we want to explores the capacity of the systems to quickly and effectively adapt to a new language. In particular, we evaluate the systems on the named entity recognition task (the simplified version of the MD task) for Dutch using the CoNLL 2002 dataset. The state-of-the-art performance for this dataset is due to Carreras et al. \shortcite{Carreras:02} in the CoNLL 2002 evaluation and Nothman et al. \shortcite{Nothman:13}. Very recently, while we were preparing this paper, Gillick el al. \shortcite{Gillick:15} introduce a multilingual language processing system and also report the performance on this dataset. Table \ref{tab:dutchCoNLL} compares the systems.


\begin{table}[htbp]
\small
\centering
\resizebox{0.39\textwidth}{!}{
\begin{tabular}{l|c|c|c}
System & P & R & F1 \\ \hline
State-of-the-art in CoNLL& 77.83 & 76.29 & 77.05 \\ \hline
Nothman et al. \shortcite{Nothman:13} & - & - & 78.60 \\ \hline
Gillick el al. \shortcite{Gillick:15} & - & - & 78.08 \\
{\it Gillick el al. \shortcite{Gillick:15}}* & - & - & {\it 82.84} \\ \hline
MEMM & 80.25 & 77.52 & 78.86 \\ \hline
BASIC & 82.98 & 81.53 & 82.25 \\ 
BIDIRECT & 84.08 & 82.82 & \textbf{83.45} \\ 
\end{tabular}
}
\caption{\small Performance on Dutch CoNLL 2002.}
\label{tab:dutchCoNLL}
\end{table}

Note that the systems in Gillick el al. \shortcite{Gillick:15} are also based on RNNs and the row labeled with * for Gillick el al. \shortcite{Gillick:15} corresponds to the system trained on multiple datasets instead of the single CoNLL dataset for Dutch, so not being comparable to ours.

The most important conclusion from the table is that the RNN models in this work significantly outperform MEMM as well as the other comparable system by large margins (up to 22\% reduction in relative error). This proves that the proposed RNN systems are less subject to the language changes than MEMM and the other systems. Finally, BIDIRECT is also significantly better than BASIC, testifying to its robustness across languages.

\section{Conclusion}

We systematically investigate various RNNs to solve the MD problem which suggests that bidirectional modeling is a very helpful mechanism for this task.   The comparison between the RNN models and the state-of-the-art systems in the literature reveals the strong promise of the RNN models. In particular, the bidirectional model achieves the best performance in the general setting (up to 9\% reduction in relative error) and outperforms a very strong baseline of the feature-based exponential models in the cross-domain setting, thus demonstrating its robustness across domains. We also show that the RNN models are more portable to new languages as they are significantly better than the best reported systems for NER in Dutch (up to 22\% reduction in relative error). In the future, we plan to apply the bidirectional modeling technique to other tasks as well as study the combination of different network architectures and resources to further improve the performance of the systems.

\section*{Acknowledgment}

We would like to thank Ralph Grishman for valuable suggestions.

\bibliographystyle{acl}
\bibliography{biblio}

\end{document}

%% file: elman_jordan.tex
\ifx\setlinejoinmode\undefined
  \newcommand{\setlinejoinmode}[1]{}
\fi
\ifx\setlinecaps\undefined
  \newcommand{\setlinecaps}[1]{}
\fi
\ifx\setfont\undefined
  \newcommand{\setfont}[2]{}
\fi
\pspicture(3.134547,-5.398852)(18.490147,-1.502022)
\psscalebox{0.175572 -0.175572}{
\newrgbcolor{dialinecolor}{0.000000 0.000000 0.000000}%
\psset{linecolor=dialinecolor}
\newrgbcolor{diafillcolor}{1.000000 1.000000 1.000000}%
\psset{fillcolor=diafillcolor}
\newrgbcolor{dialinecolor}{0.443137 0.972549 0.443137}%
\psset{linecolor=dialinecolor}
\pspolygon*(24.450000,10.050000)(24.450000,13.950000)(25.550000,13.950000)(25.550000,10.050000)
\psset{linewidth=0.100000cm}
\psset{linestyle=solid}
\psset{linestyle=solid}
\setlinejoinmode{0}
\newrgbcolor{dialinecolor}{0.000000 0.000000 0.000000}%
\psset{linecolor=dialinecolor}
\pspolygon(24.450000,10.050000)(24.450000,13.950000)(25.550000,13.950000)(25.550000,10.050000)
\setfont{Helvetica}{0.800000}
\newrgbcolor{dialinecolor}{0.000000 0.000000 0.000000}%
\psset{linecolor=dialinecolor}
\rput(25.000000,12.195000){\psscalebox{1 -1}{}}
\newrgbcolor{dialinecolor}{0.662745 0.662745 0.662745}%
\psset{linecolor=dialinecolor}
\psellipse*(24.996664,18.448282)(1.153364,1.201682)
\psset{linewidth=0.100000cm}
\psset{linestyle=solid}
\psset{linestyle=solid}
\setlinejoinmode{0}
\newrgbcolor{dialinecolor}{0.000000 0.000000 0.000000}%
\psset{linecolor=dialinecolor}
\psellipse(24.996664,18.448282)(1.153364,1.201682)
\setfont{Helvetica}{0.800000}
\newrgbcolor{dialinecolor}{0.000000 0.000000 0.000000}%
\psset{linecolor=dialinecolor}
\rput(24.996664,18.643282){\psscalebox{1 -1}{}}
\newrgbcolor{dialinecolor}{1.000000 1.000000 1.000000}%
\psset{linecolor=dialinecolor}
\pspolygon*(24.470000,22.500000)(24.470000,27.050000)(25.570000,27.050000)(25.570000,22.500000)
\psset{linewidth=0.100000cm}
\psset{linestyle=solid}
\psset{linestyle=solid}
\setlinejoinmode{0}
\newrgbcolor{dialinecolor}{0.000000 0.000000 0.000000}%
\psset{linecolor=dialinecolor}
\pspolygon(24.470000,22.500000)(24.470000,27.050000)(25.570000,27.050000)(25.570000,22.500000)
\setfont{Helvetica}{0.800000}
\newrgbcolor{dialinecolor}{0.000000 0.000000 0.000000}%
\psset{linecolor=dialinecolor}
\rput(25.020000,24.970000){\psscalebox{1 -1}{}}
\psset{linewidth=0.100000cm}
\psset{linestyle=solid}
\psset{linestyle=solid}
\setlinecaps{0}
\newrgbcolor{dialinecolor}{0.000000 0.000000 0.000000}%
\psset{linecolor=dialinecolor}
\psline(25.020000,22.500000)(25.000650,20.136751)
\psset{linestyle=solid}
\setlinejoinmode{0}
\setlinecaps{0}
\newrgbcolor{dialinecolor}{0.000000 0.000000 0.000000}%
\psset{linecolor=dialinecolor}
\pspolygon*(24.997579,19.761764)(25.251665,20.259700)(25.000650,20.136751)(24.751682,20.263794)
\newrgbcolor{dialinecolor}{0.000000 0.000000 0.000000}%
\psset{linecolor=dialinecolor}
\pspolygon(24.997579,19.761764)(25.251665,20.259700)(25.000650,20.136751)(24.751682,20.263794)
\psset{linewidth=0.100000cm}
\psset{linestyle=solid}
\psset{linestyle=solid}
\setlinecaps{0}
\newrgbcolor{dialinecolor}{0.000000 0.000000 0.000000}%
\psset{linecolor=dialinecolor}
\psline(24.996664,17.246600)(24.999507,14.436803)
\psset{linestyle=solid}
\setlinejoinmode{0}
\setlinecaps{0}
\newrgbcolor{dialinecolor}{0.000000 0.000000 0.000000}%
\psset{linecolor=dialinecolor}
\pspolygon*(24.999887,14.061803)(25.249381,14.562056)(24.999507,14.436803)(24.749381,14.561550)
\newrgbcolor{dialinecolor}{0.000000 0.000000 0.000000}%
\psset{linecolor=dialinecolor}
\pspolygon(24.999887,14.061803)(25.249381,14.562056)(24.999507,14.436803)(24.749381,14.561550)
\newrgbcolor{dialinecolor}{0.443137 0.972549 0.443137}%
\psset{linecolor=dialinecolor}
\pspolygon*(30.476700,10.100000)(30.476700,14.000000)(31.576700,14.000000)(31.576700,10.100000)
\psset{linewidth=0.100000cm}
\psset{linestyle=solid}
\psset{linestyle=solid}
\setlinejoinmode{0}
\newrgbcolor{dialinecolor}{0.000000 0.000000 0.000000}%
\psset{linecolor=dialinecolor}
\pspolygon(30.476700,10.100000)(30.476700,14.000000)(31.576700,14.000000)(31.576700,10.100000)
\setfont{Helvetica}{0.800000}
\newrgbcolor{dialinecolor}{0.000000 0.000000 0.000000}%
\psset{linecolor=dialinecolor}
\rput(31.026700,12.245000){\psscalebox{1 -1}{}}
\newrgbcolor{dialinecolor}{0.662745 0.662745 0.662745}%
\psset{linecolor=dialinecolor}
\psellipse*(31.023364,18.498282)(1.153364,1.201682)
\psset{linewidth=0.100000cm}
\psset{linestyle=solid}
\psset{linestyle=solid}
\setlinejoinmode{0}
\newrgbcolor{dialinecolor}{0.000000 0.000000 0.000000}%
\psset{linecolor=dialinecolor}
\psellipse(31.023364,18.498282)(1.153364,1.201682)
\setfont{Helvetica}{0.800000}
\newrgbcolor{dialinecolor}{0.000000 0.000000 0.000000}%
\psset{linecolor=dialinecolor}
\rput(31.023364,18.693282){\psscalebox{1 -1}{}}
\newrgbcolor{dialinecolor}{1.000000 1.000000 1.000000}%
\psset{linecolor=dialinecolor}
\pspolygon*(30.496700,22.550000)(30.496700,27.100000)(31.596700,27.100000)(31.596700,22.550000)
\psset{linewidth=0.100000cm}
\psset{linestyle=solid}
\psset{linestyle=solid}
\setlinejoinmode{0}
\newrgbcolor{dialinecolor}{0.000000 0.000000 0.000000}%
\psset{linecolor=dialinecolor}
\pspolygon(30.496700,22.550000)(30.496700,27.100000)(31.596700,27.100000)(31.596700,22.550000)
\setfont{Helvetica}{0.800000}
\newrgbcolor{dialinecolor}{0.000000 0.000000 0.000000}%
\psset{linecolor=dialinecolor}
\rput(31.046700,25.020000){\psscalebox{1 -1}{}}
\psset{linewidth=0.100000cm}
\psset{linestyle=solid}
\psset{linestyle=solid}
\setlinecaps{0}
\newrgbcolor{dialinecolor}{0.000000 0.000000 0.000000}%
\psset{linecolor=dialinecolor}
\psline(31.046700,22.550000)(31.027350,20.186751)
\psset{linestyle=solid}
\setlinejoinmode{0}
\setlinecaps{0}
\newrgbcolor{dialinecolor}{0.000000 0.000000 0.000000}%
\psset{linecolor=dialinecolor}
\pspolygon*(31.024279,19.811764)(31.278365,20.309700)(31.027350,20.186751)(30.778382,20.313794)
\newrgbcolor{dialinecolor}{0.000000 0.000000 0.000000}%
\psset{linecolor=dialinecolor}
\pspolygon(31.024279,19.811764)(31.278365,20.309700)(31.027350,20.186751)(30.778382,20.313794)
\psset{linewidth=0.100000cm}
\psset{linestyle=solid}
\psset{linestyle=solid}
\setlinecaps{0}
\newrgbcolor{dialinecolor}{0.000000 0.000000 0.000000}%
\psset{linecolor=dialinecolor}
\psline(31.023364,17.296600)(31.026207,14.486803)
\psset{linestyle=solid}
\setlinejoinmode{0}
\setlinecaps{0}
\newrgbcolor{dialinecolor}{0.000000 0.000000 0.000000}%
\psset{linecolor=dialinecolor}
\pspolygon*(31.026587,14.111803)(31.276081,14.612056)(31.026207,14.486803)(30.776081,14.611550)
\newrgbcolor{dialinecolor}{0.000000 0.000000 0.000000}%
\psset{linecolor=dialinecolor}
\pspolygon(31.026587,14.111803)(31.276081,14.612056)(31.026207,14.486803)(30.776081,14.611550)
\newrgbcolor{dialinecolor}{0.443137 0.972549 0.443137}%
\psset{linecolor=dialinecolor}
\pspolygon*(36.516700,10.100000)(36.516700,14.000000)(37.616700,14.000000)(37.616700,10.100000)
\psset{linewidth=0.100000cm}
\psset{linestyle=solid}
\psset{linestyle=solid}
\setlinejoinmode{0}
\newrgbcolor{dialinecolor}{0.000000 0.000000 0.000000}%
\psset{linecolor=dialinecolor}
\pspolygon(36.516700,10.100000)(36.516700,14.000000)(37.616700,14.000000)(37.616700,10.100000)
\setfont{Helvetica}{0.800000}
\newrgbcolor{dialinecolor}{0.000000 0.000000 0.000000}%
\psset{linecolor=dialinecolor}
\rput(37.066700,12.245000){\psscalebox{1 -1}{}}
\newrgbcolor{dialinecolor}{0.662745 0.662745 0.662745}%
\psset{linecolor=dialinecolor}
\psellipse*(37.063364,18.498282)(1.153364,1.201682)
\psset{linewidth=0.100000cm}
\psset{linestyle=solid}
\psset{linestyle=solid}
\setlinejoinmode{0}
\newrgbcolor{dialinecolor}{0.000000 0.000000 0.000000}%
\psset{linecolor=dialinecolor}
\psellipse(37.063364,18.498282)(1.153364,1.201682)
\setfont{Helvetica}{0.800000}
\newrgbcolor{dialinecolor}{0.000000 0.000000 0.000000}%
\psset{linecolor=dialinecolor}
\rput(37.063364,18.693282){\psscalebox{1 -1}{}}
\newrgbcolor{dialinecolor}{1.000000 1.000000 1.000000}%
\psset{linecolor=dialinecolor}
\pspolygon*(36.536700,22.550000)(36.536700,27.100000)(37.636700,27.100000)(37.636700,22.550000)
\psset{linewidth=0.100000cm}
\psset{linestyle=solid}
\psset{linestyle=solid}
\setlinejoinmode{0}
\newrgbcolor{dialinecolor}{0.000000 0.000000 0.000000}%
\psset{linecolor=dialinecolor}
\pspolygon(36.536700,22.550000)(36.536700,27.100000)(37.636700,27.100000)(37.636700,22.550000)
\setfont{Helvetica}{0.800000}
\newrgbcolor{dialinecolor}{0.000000 0.000000 0.000000}%
\psset{linecolor=dialinecolor}
\rput(37.086700,25.020000){\psscalebox{1 -1}{}}
\psset{linewidth=0.100000cm}
\psset{linestyle=solid}
\psset{linestyle=solid}
\setlinecaps{0}
\newrgbcolor{dialinecolor}{0.000000 0.000000 0.000000}%
\psset{linecolor=dialinecolor}
\psline(37.086700,22.550000)(37.067350,20.186751)
\psset{linestyle=solid}
\setlinejoinmode{0}
\setlinecaps{0}
\newrgbcolor{dialinecolor}{0.000000 0.000000 0.000000}%
\psset{linecolor=dialinecolor}
\pspolygon*(37.064279,19.811764)(37.318365,20.309700)(37.067350,20.186751)(36.818382,20.313794)
\newrgbcolor{dialinecolor}{0.000000 0.000000 0.000000}%
\psset{linecolor=dialinecolor}
\pspolygon(37.064279,19.811764)(37.318365,20.309700)(37.067350,20.186751)(36.818382,20.313794)
\psset{linewidth=0.100000cm}
\psset{linestyle=solid}
\psset{linestyle=solid}
\setlinecaps{0}
\newrgbcolor{dialinecolor}{0.000000 0.000000 0.000000}%
\psset{linecolor=dialinecolor}
\psline(37.063364,17.296600)(37.066207,14.486803)
\psset{linestyle=solid}
\setlinejoinmode{0}
\setlinecaps{0}
\newrgbcolor{dialinecolor}{0.000000 0.000000 0.000000}%
\psset{linecolor=dialinecolor}
\pspolygon*(37.066587,14.111803)(37.316081,14.612056)(37.066207,14.486803)(36.816081,14.611550)
\newrgbcolor{dialinecolor}{0.000000 0.000000 0.000000}%
\psset{linecolor=dialinecolor}
\pspolygon(37.066587,14.111803)(37.316081,14.612056)(37.066207,14.486803)(36.816081,14.611550)
\newrgbcolor{dialinecolor}{0.443137 0.972549 0.443137}%
\psset{linecolor=dialinecolor}
\pspolygon*(51.676700,10.100000)(51.676700,14.000000)(52.776700,14.000000)(52.776700,10.100000)
\psset{linewidth=0.100000cm}
\psset{linestyle=solid}
\psset{linestyle=solid}
\setlinejoinmode{0}
\newrgbcolor{dialinecolor}{0.000000 0.000000 0.000000}%
\psset{linecolor=dialinecolor}
\pspolygon(51.676700,10.100000)(51.676700,14.000000)(52.776700,14.000000)(52.776700,10.100000)
\setfont{Helvetica}{0.800000}
\newrgbcolor{dialinecolor}{0.000000 0.000000 0.000000}%
\psset{linecolor=dialinecolor}
\rput(52.226700,12.245000){\psscalebox{1 -1}{}}
\newrgbcolor{dialinecolor}{0.662745 0.662745 0.662745}%
\psset{linecolor=dialinecolor}
\psellipse*(52.223364,18.498282)(1.153364,1.201682)
\psset{linewidth=0.100000cm}
\psset{linestyle=solid}
\psset{linestyle=solid}
\setlinejoinmode{0}
\newrgbcolor{dialinecolor}{0.000000 0.000000 0.000000}%
\psset{linecolor=dialinecolor}
\psellipse(52.223364,18.498282)(1.153364,1.201682)
\setfont{Helvetica}{0.800000}
\newrgbcolor{dialinecolor}{0.000000 0.000000 0.000000}%
\psset{linecolor=dialinecolor}
\rput(52.223364,18.693282){\psscalebox{1 -1}{}}
\newrgbcolor{dialinecolor}{1.000000 1.000000 1.000000}%
\psset{linecolor=dialinecolor}
\pspolygon*(51.696700,22.550000)(51.696700,27.100000)(52.796700,27.100000)(52.796700,22.550000)
\psset{linewidth=0.100000cm}
\psset{linestyle=solid}
\psset{linestyle=solid}
\setlinejoinmode{0}
\newrgbcolor{dialinecolor}{0.000000 0.000000 0.000000}%
\psset{linecolor=dialinecolor}
\pspolygon(51.696700,22.550000)(51.696700,27.100000)(52.796700,27.100000)(52.796700,22.550000)
\setfont{Helvetica}{0.800000}
\newrgbcolor{dialinecolor}{0.000000 0.000000 0.000000}%
\psset{linecolor=dialinecolor}
\rput(52.246700,25.020000){\psscalebox{1 -1}{}}
\psset{linewidth=0.100000cm}
\psset{linestyle=solid}
\psset{linestyle=solid}
\setlinecaps{0}
\newrgbcolor{dialinecolor}{0.000000 0.000000 0.000000}%
\psset{linecolor=dialinecolor}
\psline(52.246700,22.550000)(52.227350,20.186751)
\psset{linestyle=solid}
\setlinejoinmode{0}
\setlinecaps{0}
\newrgbcolor{dialinecolor}{0.000000 0.000000 0.000000}%
\psset{linecolor=dialinecolor}
\pspolygon*(52.224279,19.811764)(52.478365,20.309700)(52.227350,20.186751)(51.978382,20.313794)
\newrgbcolor{dialinecolor}{0.000000 0.000000 0.000000}%
\psset{linecolor=dialinecolor}
\pspolygon(52.224279,19.811764)(52.478365,20.309700)(52.227350,20.186751)(51.978382,20.313794)
\psset{linewidth=0.100000cm}
\psset{linestyle=solid}
\psset{linestyle=solid}
\setlinecaps{0}
\newrgbcolor{dialinecolor}{0.000000 0.000000 0.000000}%
\psset{linecolor=dialinecolor}
\psline(52.223364,17.296600)(52.226207,14.486803)
\psset{linestyle=solid}
\setlinejoinmode{0}
\setlinecaps{0}
\newrgbcolor{dialinecolor}{0.000000 0.000000 0.000000}%
\psset{linecolor=dialinecolor}
\pspolygon*(52.226587,14.111803)(52.476081,14.612056)(52.226207,14.486803)(51.976081,14.611550)
\newrgbcolor{dialinecolor}{0.000000 0.000000 0.000000}%
\psset{linecolor=dialinecolor}
\pspolygon(52.226587,14.111803)(52.476081,14.612056)(52.226207,14.486803)(51.976081,14.611550)
\newrgbcolor{dialinecolor}{0.443137 0.972549 0.443137}%
\psset{linecolor=dialinecolor}
\pspolygon*(57.696700,10.050000)(57.696700,13.950000)(58.796700,13.950000)(58.796700,10.050000)
\psset{linewidth=0.100000cm}
\psset{linestyle=solid}
\psset{linestyle=solid}
\setlinejoinmode{0}
\newrgbcolor{dialinecolor}{0.000000 0.000000 0.000000}%
\psset{linecolor=dialinecolor}
\pspolygon(57.696700,10.050000)(57.696700,13.950000)(58.796700,13.950000)(58.796700,10.050000)
\setfont{Helvetica}{0.800000}
\newrgbcolor{dialinecolor}{0.000000 0.000000 0.000000}%
\psset{linecolor=dialinecolor}
\rput(58.246700,12.195000){\psscalebox{1 -1}{}}
\newrgbcolor{dialinecolor}{0.662745 0.662745 0.662745}%
\psset{linecolor=dialinecolor}
\psellipse*(58.243364,18.448282)(1.153364,1.201682)
\psset{linewidth=0.100000cm}
\psset{linestyle=solid}
\psset{linestyle=solid}
\setlinejoinmode{0}
\newrgbcolor{dialinecolor}{0.000000 0.000000 0.000000}%
\psset{linecolor=dialinecolor}
\psellipse(58.243364,18.448282)(1.153364,1.201682)
\setfont{Helvetica}{0.800000}
\newrgbcolor{dialinecolor}{0.000000 0.000000 0.000000}%
\psset{linecolor=dialinecolor}
\rput(58.243364,18.643282){\psscalebox{1 -1}{}}
\newrgbcolor{dialinecolor}{1.000000 1.000000 1.000000}%
\psset{linecolor=dialinecolor}
\pspolygon*(57.716700,22.500000)(57.716700,27.050000)(58.816700,27.050000)(58.816700,22.500000)
\psset{linewidth=0.100000cm}
\psset{linestyle=solid}
\psset{linestyle=solid}
\setlinejoinmode{0}
\newrgbcolor{dialinecolor}{0.000000 0.000000 0.000000}%
\psset{linecolor=dialinecolor}
\pspolygon(57.716700,22.500000)(57.716700,27.050000)(58.816700,27.050000)(58.816700,22.500000)
\setfont{Helvetica}{0.800000}
\newrgbcolor{dialinecolor}{0.000000 0.000000 0.000000}%
\psset{linecolor=dialinecolor}
\rput(58.266700,24.970000){\psscalebox{1 -1}{}}
\psset{linewidth=0.100000cm}
\psset{linestyle=solid}
\psset{linestyle=solid}
\setlinecaps{0}
\newrgbcolor{dialinecolor}{0.000000 0.000000 0.000000}%
\psset{linecolor=dialinecolor}
\psline(58.266700,22.500000)(58.247350,20.136751)
\psset{linestyle=solid}
\setlinejoinmode{0}
\setlinecaps{0}
\newrgbcolor{dialinecolor}{0.000000 0.000000 0.000000}%
\psset{linecolor=dialinecolor}
\pspolygon*(58.244279,19.761764)(58.498365,20.259700)(58.247350,20.136751)(57.998382,20.263794)
\newrgbcolor{dialinecolor}{0.000000 0.000000 0.000000}%
\psset{linecolor=dialinecolor}
\pspolygon(58.244279,19.761764)(58.498365,20.259700)(58.247350,20.136751)(57.998382,20.263794)
\psset{linewidth=0.100000cm}
\psset{linestyle=solid}
\psset{linestyle=solid}
\setlinecaps{0}
\newrgbcolor{dialinecolor}{0.000000 0.000000 0.000000}%
\psset{linecolor=dialinecolor}
\psline(58.243364,17.246600)(58.246207,14.436803)
\psset{linestyle=solid}
\setlinejoinmode{0}
\setlinecaps{0}
\newrgbcolor{dialinecolor}{0.000000 0.000000 0.000000}%
\psset{linecolor=dialinecolor}
\pspolygon*(58.246587,14.061803)(58.496081,14.562056)(58.246207,14.436803)(57.996081,14.561550)
\newrgbcolor{dialinecolor}{0.000000 0.000000 0.000000}%
\psset{linecolor=dialinecolor}
\pspolygon(58.246587,14.061803)(58.496081,14.562056)(58.246207,14.436803)(57.996081,14.561550)
\psset{linewidth=0.100000cm}
\psset{linestyle=solid}
\psset{linestyle=solid}
\setlinecaps{0}
\newrgbcolor{dialinecolor}{0.000000 0.000000 0.000000}%
\psset{linecolor=dialinecolor}
\psline(26.150028,18.448282)(29.383241,18.491739)
\psset{linestyle=solid}
\setlinejoinmode{0}
\setlinecaps{0}
\newrgbcolor{dialinecolor}{0.000000 0.000000 0.000000}%
\psset{linecolor=dialinecolor}
\pspolygon*(29.758207,18.496779)(29.254892,18.740037)(29.383241,18.491739)(29.261612,18.240082)
\newrgbcolor{dialinecolor}{0.000000 0.000000 0.000000}%
\psset{linecolor=dialinecolor}
\pspolygon(29.758207,18.496779)(29.254892,18.740037)(29.383241,18.491739)(29.261612,18.240082)
\psset{linewidth=0.100000cm}
\psset{linestyle=solid}
\psset{linestyle=solid}
\setlinecaps{0}
\newrgbcolor{dialinecolor}{0.000000 0.000000 0.000000}%
\psset{linecolor=dialinecolor}
\psline(32.176728,18.498282)(35.423197,18.498282)
\psset{linestyle=solid}
\setlinejoinmode{0}
\setlinecaps{0}
\newrgbcolor{dialinecolor}{0.000000 0.000000 0.000000}%
\psset{linecolor=dialinecolor}
\pspolygon*(35.798197,18.498282)(35.298197,18.748282)(35.423197,18.498282)(35.298197,18.248282)
\newrgbcolor{dialinecolor}{0.000000 0.000000 0.000000}%
\psset{linecolor=dialinecolor}
\pspolygon(35.798197,18.498282)(35.298197,18.748282)(35.423197,18.498282)(35.298197,18.248282)
\psset{linewidth=0.100000cm}
\psset{linestyle=solid}
\psset{linestyle=solid}
\setlinecaps{0}
\newrgbcolor{dialinecolor}{0.000000 0.000000 0.000000}%
\psset{linecolor=dialinecolor}
\psline(53.376728,18.498282)(56.603241,18.454836)
\psset{linestyle=solid}
\setlinejoinmode{0}
\setlinecaps{0}
\newrgbcolor{dialinecolor}{0.000000 0.000000 0.000000}%
\psset{linecolor=dialinecolor}
\pspolygon*(56.978207,18.449787)(56.481618,18.706497)(56.603241,18.454836)(56.474886,18.206542)
\newrgbcolor{dialinecolor}{0.000000 0.000000 0.000000}%
\psset{linecolor=dialinecolor}
\pspolygon(56.978207,18.449787)(56.481618,18.706497)(56.603241,18.454836)(56.474886,18.206542)
\newrgbcolor{dialinecolor}{0.662745 0.662745 0.662745}%
\psset{linecolor=dialinecolor}
\psellipse*(19.056664,18.448282)(1.153364,1.201682)
\psset{linewidth=0.100000cm}
\psset{linestyle=solid}
\psset{linestyle=solid}
\setlinejoinmode{0}
\newrgbcolor{dialinecolor}{0.000000 0.000000 0.000000}%
\psset{linecolor=dialinecolor}
\psellipse(19.056664,18.448282)(1.153364,1.201682)
\setfont{Helvetica}{0.800000}
\newrgbcolor{dialinecolor}{0.000000 0.000000 0.000000}%
\psset{linecolor=dialinecolor}
\rput(19.056664,18.643282){\psscalebox{1 -1}{}}
\psset{linewidth=0.100000cm}
\psset{linestyle=solid}
\psset{linestyle=solid}
\setlinecaps{0}
\newrgbcolor{dialinecolor}{0.000000 0.000000 0.000000}%
\psset{linecolor=dialinecolor}
\psline(20.210028,18.448282)(23.356497,18.448282)
\psset{linestyle=solid}
\setlinejoinmode{0}
\setlinecaps{0}
\newrgbcolor{dialinecolor}{0.000000 0.000000 0.000000}%
\psset{linecolor=dialinecolor}
\pspolygon*(23.731497,18.448282)(23.231497,18.698282)(23.356497,18.448282)(23.231497,18.198282)
\newrgbcolor{dialinecolor}{0.000000 0.000000 0.000000}%
\psset{linecolor=dialinecolor}
\pspolygon(23.731497,18.448282)(23.231497,18.698282)(23.356497,18.448282)(23.231497,18.198282)
\psset{linewidth=0.100000cm}
\psset{linestyle=solid}
\psset{linestyle=solid}
\setlinecaps{0}
\newrgbcolor{dialinecolor}{0.000000 0.000000 0.000000}%
\psset{linecolor=dialinecolor}
\psline(38.216728,18.498282)(41.566500,18.510085)
\psset{linestyle=solid}
\setlinejoinmode{0}
\setlinecaps{0}
\newrgbcolor{dialinecolor}{0.000000 0.000000 0.000000}%
\psset{linecolor=dialinecolor}
\pspolygon*(41.941497,18.511406)(41.440620,18.759643)(41.566500,18.510085)(41.442381,18.259646)
\newrgbcolor{dialinecolor}{0.000000 0.000000 0.000000}%
\psset{linecolor=dialinecolor}
\pspolygon(41.941497,18.511406)(41.440620,18.759643)(41.566500,18.510085)(41.442381,18.259646)
\psset{linewidth=0.200000cm}
\psset{linestyle=dotted,dotsep=0.200000}
\psset{linestyle=dotted,dotsep=0.200000}
\setlinecaps{0}
\newrgbcolor{dialinecolor}{0.000000 0.000000 0.000000}%
\psset{linecolor=dialinecolor}
\psline(42.000000,19.800000)(47.563600,19.800000)
\psset{linewidth=0.100000cm}
\psset{linestyle=solid}
\psset{linestyle=solid}
\setlinecaps{0}
\newrgbcolor{dialinecolor}{0.000000 0.000000 0.000000}%
\psset{linecolor=dialinecolor}
\psline(47.100000,18.450000)(50.583233,18.492362)
\psset{linestyle=solid}
\setlinejoinmode{0}
\setlinecaps{0}
\newrgbcolor{dialinecolor}{0.000000 0.000000 0.000000}%
\psset{linecolor=dialinecolor}
\pspolygon*(50.958205,18.496922)(50.455202,18.740823)(50.583233,18.492362)(50.461282,18.240860)
\newrgbcolor{dialinecolor}{0.000000 0.000000 0.000000}%
\psset{linecolor=dialinecolor}
\pspolygon(50.958205,18.496922)(50.455202,18.740823)(50.583233,18.492362)(50.461282,18.240860)
\newrgbcolor{dialinecolor}{0.443137 0.972549 0.443137}%
\psset{linecolor=dialinecolor}
\pspolygon*(72.566772,10.025018)(72.566772,13.925018)(73.666772,13.925018)(73.666772,10.025018)
\psset{linewidth=0.100000cm}
\psset{linestyle=solid}
\psset{linestyle=solid}
\setlinejoinmode{0}
\newrgbcolor{dialinecolor}{0.000000 0.000000 0.000000}%
\psset{linecolor=dialinecolor}
\pspolygon(72.566772,10.025018)(72.566772,13.925018)(73.666772,13.925018)(73.666772,10.025018)
\setfont{Helvetica}{0.800000}
\newrgbcolor{dialinecolor}{0.000000 0.000000 0.000000}%
\psset{linecolor=dialinecolor}
\rput(73.116772,12.170018){\psscalebox{1 -1}{}}
\newrgbcolor{dialinecolor}{0.662745 0.662745 0.662745}%
\psset{linecolor=dialinecolor}
\psellipse*(73.113436,18.423300)(1.153364,1.201682)
\psset{linewidth=0.100000cm}
\psset{linestyle=solid}
\psset{linestyle=solid}
\setlinejoinmode{0}
\newrgbcolor{dialinecolor}{0.000000 0.000000 0.000000}%
\psset{linecolor=dialinecolor}
\psellipse(73.113436,18.423300)(1.153364,1.201682)
\setfont{Helvetica}{0.800000}
\newrgbcolor{dialinecolor}{0.000000 0.000000 0.000000}%
\psset{linecolor=dialinecolor}
\rput(73.113436,18.618300){\psscalebox{1 -1}{}}
\newrgbcolor{dialinecolor}{1.000000 1.000000 1.000000}%
\psset{linecolor=dialinecolor}
\pspolygon*(72.586772,22.475018)(72.586772,27.025018)(73.686772,27.025018)(73.686772,22.475018)
\psset{linewidth=0.100000cm}
\psset{linestyle=solid}
\psset{linestyle=solid}
\setlinejoinmode{0}
\newrgbcolor{dialinecolor}{0.000000 0.000000 0.000000}%
\psset{linecolor=dialinecolor}
\pspolygon(72.586772,22.475018)(72.586772,27.025018)(73.686772,27.025018)(73.686772,22.475018)
\setfont{Helvetica}{0.800000}
\newrgbcolor{dialinecolor}{0.000000 0.000000 0.000000}%
\psset{linecolor=dialinecolor}
\rput(73.136772,24.945018){\psscalebox{1 -1}{}}
\psset{linewidth=0.100000cm}
\psset{linestyle=solid}
\psset{linestyle=solid}
\setlinecaps{0}
\newrgbcolor{dialinecolor}{0.000000 0.000000 0.000000}%
\psset{linecolor=dialinecolor}
\psline(73.136772,22.475018)(73.117422,20.111769)
\psset{linestyle=solid}
\setlinejoinmode{0}
\setlinecaps{0}
\newrgbcolor{dialinecolor}{0.000000 0.000000 0.000000}%
\psset{linecolor=dialinecolor}
\pspolygon*(73.114352,19.736782)(73.368437,20.234718)(73.117422,20.111769)(72.868454,20.238812)
\newrgbcolor{dialinecolor}{0.000000 0.000000 0.000000}%
\psset{linecolor=dialinecolor}
\pspolygon(73.114352,19.736782)(73.368437,20.234718)(73.117422,20.111769)(72.868454,20.238812)
\psset{linewidth=0.100000cm}
\psset{linestyle=solid}
\psset{linestyle=solid}
\setlinecaps{0}
\newrgbcolor{dialinecolor}{0.000000 0.000000 0.000000}%
\psset{linecolor=dialinecolor}
\psline(73.113436,17.221618)(73.116280,14.411821)
\psset{linestyle=solid}
\setlinejoinmode{0}
\setlinecaps{0}
\newrgbcolor{dialinecolor}{0.000000 0.000000 0.000000}%
\psset{linecolor=dialinecolor}
\pspolygon*(73.116659,14.036821)(73.366153,14.537074)(73.116280,14.411821)(72.866153,14.536568)
\newrgbcolor{dialinecolor}{0.000000 0.000000 0.000000}%
\psset{linecolor=dialinecolor}
\pspolygon(73.116659,14.036821)(73.366153,14.537074)(73.116280,14.411821)(72.866153,14.536568)
\newrgbcolor{dialinecolor}{0.443137 0.972549 0.443137}%
\psset{linecolor=dialinecolor}
\pspolygon*(78.593472,10.075018)(78.593472,13.975018)(79.693472,13.975018)(79.693472,10.075018)
\psset{linewidth=0.100000cm}
\psset{linestyle=solid}
\psset{linestyle=solid}
\setlinejoinmode{0}
\newrgbcolor{dialinecolor}{0.000000 0.000000 0.000000}%
\psset{linecolor=dialinecolor}
\pspolygon(78.593472,10.075018)(78.593472,13.975018)(79.693472,13.975018)(79.693472,10.075018)
\setfont{Helvetica}{0.800000}
\newrgbcolor{dialinecolor}{0.000000 0.000000 0.000000}%
\psset{linecolor=dialinecolor}
\rput(79.143472,12.220018){\psscalebox{1 -1}{}}
\newrgbcolor{dialinecolor}{0.662745 0.662745 0.662745}%
\psset{linecolor=dialinecolor}
\psellipse*(79.140136,18.473300)(1.153364,1.201682)
\psset{linewidth=0.100000cm}
\psset{linestyle=solid}
\psset{linestyle=solid}
\setlinejoinmode{0}
\newrgbcolor{dialinecolor}{0.000000 0.000000 0.000000}%
\psset{linecolor=dialinecolor}
\psellipse(79.140136,18.473300)(1.153364,1.201682)
\setfont{Helvetica}{0.800000}
\newrgbcolor{dialinecolor}{0.000000 0.000000 0.000000}%
\psset{linecolor=dialinecolor}
\rput(79.140136,18.668300){\psscalebox{1 -1}{}}
\newrgbcolor{dialinecolor}{1.000000 1.000000 1.000000}%
\psset{linecolor=dialinecolor}
\pspolygon*(78.613472,22.525018)(78.613472,27.075018)(79.713472,27.075018)(79.713472,22.525018)
\psset{linewidth=0.100000cm}
\psset{linestyle=solid}
\psset{linestyle=solid}
\setlinejoinmode{0}
\newrgbcolor{dialinecolor}{0.000000 0.000000 0.000000}%
\psset{linecolor=dialinecolor}
\pspolygon(78.613472,22.525018)(78.613472,27.075018)(79.713472,27.075018)(79.713472,22.525018)
\setfont{Helvetica}{0.800000}
\newrgbcolor{dialinecolor}{0.000000 0.000000 0.000000}%
\psset{linecolor=dialinecolor}
\rput(79.163472,24.995018){\psscalebox{1 -1}{}}
\psset{linewidth=0.100000cm}
\psset{linestyle=solid}
\psset{linestyle=solid}
\setlinecaps{0}
\newrgbcolor{dialinecolor}{0.000000 0.000000 0.000000}%
\psset{linecolor=dialinecolor}
\psline(79.163472,22.525018)(79.144122,20.161769)
\psset{linestyle=solid}
\setlinejoinmode{0}
\setlinecaps{0}
\newrgbcolor{dialinecolor}{0.000000 0.000000 0.000000}%
\psset{linecolor=dialinecolor}
\pspolygon*(79.141052,19.786782)(79.395137,20.284718)(79.144122,20.161769)(78.895154,20.288812)
\newrgbcolor{dialinecolor}{0.000000 0.000000 0.000000}%
\psset{linecolor=dialinecolor}
\pspolygon(79.141052,19.786782)(79.395137,20.284718)(79.144122,20.161769)(78.895154,20.288812)
\psset{linewidth=0.100000cm}
\psset{linestyle=solid}
\psset{linestyle=solid}
\setlinecaps{0}
\newrgbcolor{dialinecolor}{0.000000 0.000000 0.000000}%
\psset{linecolor=dialinecolor}
\psline(79.140136,17.271618)(79.142980,14.461821)
\psset{linestyle=solid}
\setlinejoinmode{0}
\setlinecaps{0}
\newrgbcolor{dialinecolor}{0.000000 0.000000 0.000000}%
\psset{linecolor=dialinecolor}
\pspolygon*(79.143359,14.086821)(79.392853,14.587074)(79.142980,14.461821)(78.892853,14.586568)
\newrgbcolor{dialinecolor}{0.000000 0.000000 0.000000}%
\psset{linecolor=dialinecolor}
\pspolygon(79.143359,14.086821)(79.392853,14.587074)(79.142980,14.461821)(78.892853,14.586568)
\newrgbcolor{dialinecolor}{0.443137 0.972549 0.443137}%
\psset{linecolor=dialinecolor}
\pspolygon*(84.633472,10.075018)(84.633472,13.975018)(85.733472,13.975018)(85.733472,10.075018)
\psset{linewidth=0.100000cm}
\psset{linestyle=solid}
\psset{linestyle=solid}
\setlinejoinmode{0}
\newrgbcolor{dialinecolor}{0.000000 0.000000 0.000000}%
\psset{linecolor=dialinecolor}
\pspolygon(84.633472,10.075018)(84.633472,13.975018)(85.733472,13.975018)(85.733472,10.075018)
\setfont{Helvetica}{0.800000}
\newrgbcolor{dialinecolor}{0.000000 0.000000 0.000000}%
\psset{linecolor=dialinecolor}
\rput(85.183472,12.220018){\psscalebox{1 -1}{}}
\newrgbcolor{dialinecolor}{0.662745 0.662745 0.662745}%
\psset{linecolor=dialinecolor}
\psellipse*(85.180136,18.473300)(1.153364,1.201682)
\psset{linewidth=0.100000cm}
\psset{linestyle=solid}
\psset{linestyle=solid}
\setlinejoinmode{0}
\newrgbcolor{dialinecolor}{0.000000 0.000000 0.000000}%
\psset{linecolor=dialinecolor}
\psellipse(85.180136,18.473300)(1.153364,1.201682)
\setfont{Helvetica}{0.800000}
\newrgbcolor{dialinecolor}{0.000000 0.000000 0.000000}%
\psset{linecolor=dialinecolor}
\rput(85.180136,18.668300){\psscalebox{1 -1}{}}
\newrgbcolor{dialinecolor}{1.000000 1.000000 1.000000}%
\psset{linecolor=dialinecolor}
\pspolygon*(84.653472,22.525018)(84.653472,27.075018)(85.753472,27.075018)(85.753472,22.525018)
\psset{linewidth=0.100000cm}
\psset{linestyle=solid}
\psset{linestyle=solid}
\setlinejoinmode{0}
\newrgbcolor{dialinecolor}{0.000000 0.000000 0.000000}%
\psset{linecolor=dialinecolor}
\pspolygon(84.653472,22.525018)(84.653472,27.075018)(85.753472,27.075018)(85.753472,22.525018)
\setfont{Helvetica}{0.800000}
\newrgbcolor{dialinecolor}{0.000000 0.000000 0.000000}%
\psset{linecolor=dialinecolor}
\rput(85.203472,24.995018){\psscalebox{1 -1}{}}
\psset{linewidth=0.100000cm}
\psset{linestyle=solid}
\psset{linestyle=solid}
\setlinecaps{0}
\newrgbcolor{dialinecolor}{0.000000 0.000000 0.000000}%
\psset{linecolor=dialinecolor}
\psline(85.203472,22.525018)(85.184122,20.161769)
\psset{linestyle=solid}
\setlinejoinmode{0}
\setlinecaps{0}
\newrgbcolor{dialinecolor}{0.000000 0.000000 0.000000}%
\psset{linecolor=dialinecolor}
\pspolygon*(85.181052,19.786782)(85.435137,20.284718)(85.184122,20.161769)(84.935154,20.288812)
\newrgbcolor{dialinecolor}{0.000000 0.000000 0.000000}%
\psset{linecolor=dialinecolor}
\pspolygon(85.181052,19.786782)(85.435137,20.284718)(85.184122,20.161769)(84.935154,20.288812)
\psset{linewidth=0.100000cm}
\psset{linestyle=solid}
\psset{linestyle=solid}
\setlinecaps{0}
\newrgbcolor{dialinecolor}{0.000000 0.000000 0.000000}%
\psset{linecolor=dialinecolor}
\psline(85.180136,17.271618)(85.182980,14.461821)
\psset{linestyle=solid}
\setlinejoinmode{0}
\setlinecaps{0}
\newrgbcolor{dialinecolor}{0.000000 0.000000 0.000000}%
\psset{linecolor=dialinecolor}
\pspolygon*(85.183359,14.086821)(85.432853,14.587074)(85.182980,14.461821)(84.932853,14.586568)
\newrgbcolor{dialinecolor}{0.000000 0.000000 0.000000}%
\psset{linecolor=dialinecolor}
\pspolygon(85.183359,14.086821)(85.432853,14.587074)(85.182980,14.461821)(84.932853,14.586568)
\newrgbcolor{dialinecolor}{0.443137 0.972549 0.443137}%
\psset{linecolor=dialinecolor}
\pspolygon*(97.543472,10.075018)(97.543472,13.975018)(98.643472,13.975018)(98.643472,10.075018)
\psset{linewidth=0.100000cm}
\psset{linestyle=solid}
\psset{linestyle=solid}
\setlinejoinmode{0}
\newrgbcolor{dialinecolor}{0.000000 0.000000 0.000000}%
\psset{linecolor=dialinecolor}
\pspolygon(97.543472,10.075018)(97.543472,13.975018)(98.643472,13.975018)(98.643472,10.075018)
\setfont{Helvetica}{0.800000}
\newrgbcolor{dialinecolor}{0.000000 0.000000 0.000000}%
\psset{linecolor=dialinecolor}
\rput(98.093472,12.220018){\psscalebox{1 -1}{}}
\newrgbcolor{dialinecolor}{0.662745 0.662745 0.662745}%
\psset{linecolor=dialinecolor}
\psellipse*(98.090136,18.473300)(1.153364,1.201682)
\psset{linewidth=0.100000cm}
\psset{linestyle=solid}
\psset{linestyle=solid}
\setlinejoinmode{0}
\newrgbcolor{dialinecolor}{0.000000 0.000000 0.000000}%
\psset{linecolor=dialinecolor}
\psellipse(98.090136,18.473300)(1.153364,1.201682)
\setfont{Helvetica}{0.800000}
\newrgbcolor{dialinecolor}{0.000000 0.000000 0.000000}%
\psset{linecolor=dialinecolor}
\rput(98.090136,18.668300){\psscalebox{1 -1}{}}
\newrgbcolor{dialinecolor}{1.000000 1.000000 1.000000}%
\psset{linecolor=dialinecolor}
\pspolygon*(97.563472,22.525018)(97.563472,27.075018)(98.663472,27.075018)(98.663472,22.525018)
\psset{linewidth=0.100000cm}
\psset{linestyle=solid}
\psset{linestyle=solid}
\setlinejoinmode{0}
\newrgbcolor{dialinecolor}{0.000000 0.000000 0.000000}%
\psset{linecolor=dialinecolor}
\pspolygon(97.563472,22.525018)(97.563472,27.075018)(98.663472,27.075018)(98.663472,22.525018)
\setfont{Helvetica}{0.800000}
\newrgbcolor{dialinecolor}{0.000000 0.000000 0.000000}%
\psset{linecolor=dialinecolor}
\rput(98.113472,24.995018){\psscalebox{1 -1}{}}
\psset{linewidth=0.100000cm}
\psset{linestyle=solid}
\psset{linestyle=solid}
\setlinecaps{0}
\newrgbcolor{dialinecolor}{0.000000 0.000000 0.000000}%
\psset{linecolor=dialinecolor}
\psline(98.113472,22.525018)(98.094122,20.161769)
\psset{linestyle=solid}
\setlinejoinmode{0}
\setlinecaps{0}
\newrgbcolor{dialinecolor}{0.000000 0.000000 0.000000}%
\psset{linecolor=dialinecolor}
\pspolygon*(98.091052,19.786782)(98.345137,20.284718)(98.094122,20.161769)(97.845154,20.288812)
\newrgbcolor{dialinecolor}{0.000000 0.000000 0.000000}%
\psset{linecolor=dialinecolor}
\pspolygon(98.091052,19.786782)(98.345137,20.284718)(98.094122,20.161769)(97.845154,20.288812)
\psset{linewidth=0.100000cm}
\psset{linestyle=solid}
\psset{linestyle=solid}
\setlinecaps{0}
\newrgbcolor{dialinecolor}{0.000000 0.000000 0.000000}%
\psset{linecolor=dialinecolor}
\psline(98.090136,17.271618)(98.092980,14.461821)
\psset{linestyle=solid}
\setlinejoinmode{0}
\setlinecaps{0}
\newrgbcolor{dialinecolor}{0.000000 0.000000 0.000000}%
\psset{linecolor=dialinecolor}
\pspolygon*(98.093359,14.086821)(98.342853,14.587074)(98.092980,14.461821)(97.842853,14.586568)
\newrgbcolor{dialinecolor}{0.000000 0.000000 0.000000}%
\psset{linecolor=dialinecolor}
\pspolygon(98.093359,14.086821)(98.342853,14.587074)(98.092980,14.461821)(97.842853,14.586568)
\newrgbcolor{dialinecolor}{0.443137 0.972549 0.443137}%
\psset{linecolor=dialinecolor}
\pspolygon*(103.563472,10.025018)(103.563472,13.925018)(104.663472,13.925018)(104.663472,10.025018)
\psset{linewidth=0.100000cm}
\psset{linestyle=solid}
\psset{linestyle=solid}
\setlinejoinmode{0}
\newrgbcolor{dialinecolor}{0.000000 0.000000 0.000000}%
\psset{linecolor=dialinecolor}
\pspolygon(103.563472,10.025018)(103.563472,13.925018)(104.663472,13.925018)(104.663472,10.025018)
\setfont{Helvetica}{0.800000}
\newrgbcolor{dialinecolor}{0.000000 0.000000 0.000000}%
\psset{linecolor=dialinecolor}
\rput(104.113472,12.170018){\psscalebox{1 -1}{}}
\newrgbcolor{dialinecolor}{0.662745 0.662745 0.662745}%
\psset{linecolor=dialinecolor}
\psellipse*(104.110136,18.423300)(1.153364,1.201682)
\psset{linewidth=0.100000cm}
\psset{linestyle=solid}
\psset{linestyle=solid}
\setlinejoinmode{0}
\newrgbcolor{dialinecolor}{0.000000 0.000000 0.000000}%
\psset{linecolor=dialinecolor}
\psellipse(104.110136,18.423300)(1.153364,1.201682)
\setfont{Helvetica}{0.800000}
\newrgbcolor{dialinecolor}{0.000000 0.000000 0.000000}%
\psset{linecolor=dialinecolor}
\rput(104.110136,18.618300){\psscalebox{1 -1}{}}
\newrgbcolor{dialinecolor}{1.000000 1.000000 1.000000}%
\psset{linecolor=dialinecolor}
\pspolygon*(103.583472,22.475018)(103.583472,27.025018)(104.683472,27.025018)(104.683472,22.475018)
\psset{linewidth=0.100000cm}
\psset{linestyle=solid}
\psset{linestyle=solid}
\setlinejoinmode{0}
\newrgbcolor{dialinecolor}{0.000000 0.000000 0.000000}%
\psset{linecolor=dialinecolor}
\pspolygon(103.583472,22.475018)(103.583472,27.025018)(104.683472,27.025018)(104.683472,22.475018)
\setfont{Helvetica}{0.800000}
\newrgbcolor{dialinecolor}{0.000000 0.000000 0.000000}%
\psset{linecolor=dialinecolor}
\rput(104.133472,24.945018){\psscalebox{1 -1}{}}
\psset{linewidth=0.100000cm}
\psset{linestyle=solid}
\psset{linestyle=solid}
\setlinecaps{0}
\newrgbcolor{dialinecolor}{0.000000 0.000000 0.000000}%
\psset{linecolor=dialinecolor}
\psline(104.133472,22.475018)(104.114122,20.111769)
\psset{linestyle=solid}
\setlinejoinmode{0}
\setlinecaps{0}
\newrgbcolor{dialinecolor}{0.000000 0.000000 0.000000}%
\psset{linecolor=dialinecolor}
\pspolygon*(104.111052,19.736782)(104.365137,20.234718)(104.114122,20.111769)(103.865154,20.238812)
\newrgbcolor{dialinecolor}{0.000000 0.000000 0.000000}%
\psset{linecolor=dialinecolor}
\pspolygon(104.111052,19.736782)(104.365137,20.234718)(104.114122,20.111769)(103.865154,20.238812)
\psset{linewidth=0.100000cm}
\psset{linestyle=solid}
\psset{linestyle=solid}
\setlinecaps{0}
\newrgbcolor{dialinecolor}{0.000000 0.000000 0.000000}%
\psset{linecolor=dialinecolor}
\psline(104.110136,17.221618)(104.112980,14.411821)
\psset{linestyle=solid}
\setlinejoinmode{0}
\setlinecaps{0}
\newrgbcolor{dialinecolor}{0.000000 0.000000 0.000000}%
\psset{linecolor=dialinecolor}
\pspolygon*(104.113359,14.036821)(104.362853,14.537074)(104.112980,14.411821)(103.862853,14.536568)
\newrgbcolor{dialinecolor}{0.000000 0.000000 0.000000}%
\psset{linecolor=dialinecolor}
\pspolygon(104.113359,14.036821)(104.362853,14.537074)(104.112980,14.411821)(103.862853,14.536568)
\psset{linewidth=0.200000cm}
\psset{linestyle=dotted,dotsep=0.200000}
\psset{linestyle=dotted,dotsep=0.200000}
\setlinecaps{0}
\newrgbcolor{dialinecolor}{0.000000 0.000000 0.000000}%
\psset{linecolor=dialinecolor}
\psline(90.766772,19.775018)(95.813600,19.750000)
\psset{linewidth=0.100000cm}
\psset{linestyle=solid}
\psset{linestyle=solid}
\setlinecaps{0}
\newrgbcolor{dialinecolor}{0.000000 0.000000 0.000000}%
\psset{linecolor=dialinecolor}
\psline(73.666772,13.925018)(77.943352,17.320863)
\psset{linestyle=solid}
\setlinejoinmode{0}
\setlinecaps{0}
\newrgbcolor{dialinecolor}{0.000000 0.000000 0.000000}%
\psset{linecolor=dialinecolor}
\pspolygon*(78.237028,17.554057)(77.689997,17.438915)(77.943352,17.320863)(78.000924,17.047347)
\newrgbcolor{dialinecolor}{0.000000 0.000000 0.000000}%
\psset{linecolor=dialinecolor}
\pspolygon(78.237028,17.554057)(77.689997,17.438915)(77.943352,17.320863)(78.000924,17.047347)
\psset{linewidth=0.100000cm}
\psset{linestyle=solid}
\psset{linestyle=solid}
\setlinecaps{0}
\newrgbcolor{dialinecolor}{0.000000 0.000000 0.000000}%
\psset{linecolor=dialinecolor}
\psline(79.693472,13.975018)(83.980942,17.323923)
\psset{linestyle=solid}
\setlinejoinmode{0}
\setlinecaps{0}
\newrgbcolor{dialinecolor}{0.000000 0.000000 0.000000}%
\psset{linecolor=dialinecolor}
\pspolygon*(84.276474,17.554760)(83.728540,17.443998)(83.980942,17.323923)(84.036323,17.049956)
\newrgbcolor{dialinecolor}{0.000000 0.000000 0.000000}%
\psset{linecolor=dialinecolor}
\pspolygon(84.276474,17.554760)(83.728540,17.443998)(83.980942,17.323923)(84.036323,17.049956)
\psset{linewidth=0.100000cm}
\psset{linestyle=solid}
\psset{linestyle=solid}
\setlinecaps{0}
\newrgbcolor{dialinecolor}{0.000000 0.000000 0.000000}%
\psset{linecolor=dialinecolor}
\psline(85.458472,13.975018)(90.229798,17.409361)
\psset{linestyle=solid}
\setlinejoinmode{0}
\setlinecaps{0}
\newrgbcolor{dialinecolor}{0.000000 0.000000 0.000000}%
\psset{linecolor=dialinecolor}
\pspolygon*(90.534154,17.628432)(89.982298,17.539241)(90.229798,17.409361)(90.274393,17.133433)
\newrgbcolor{dialinecolor}{0.000000 0.000000 0.000000}%
\psset{linecolor=dialinecolor}
\pspolygon(90.534154,17.628432)(89.982298,17.539241)(90.229798,17.409361)(90.274393,17.133433)
\psset{linewidth=0.100000cm}
\psset{linestyle=solid}
\psset{linestyle=solid}
\setlinecaps{0}
\newrgbcolor{dialinecolor}{0.000000 0.000000 0.000000}%
\psset{linecolor=dialinecolor}
\psline(98.643472,13.975018)(102.909566,17.275693)
\psset{linestyle=solid}
\setlinejoinmode{0}
\setlinecaps{0}
\newrgbcolor{dialinecolor}{0.000000 0.000000 0.000000}%
\psset{linecolor=dialinecolor}
\pspolygon*(103.206158,17.505167)(102.657719,17.396931)(102.909566,17.275693)(102.963684,17.001474)
\newrgbcolor{dialinecolor}{0.000000 0.000000 0.000000}%
\psset{linecolor=dialinecolor}
\pspolygon(103.206158,17.505167)(102.657719,17.396931)(102.909566,17.275693)(102.963684,17.001474)
\newrgbcolor{dialinecolor}{0.443137 0.972549 0.443137}%
\psset{linecolor=dialinecolor}
\pspolygon*(66.490400,10.075000)(66.490400,13.975000)(67.590400,13.975000)(67.590400,10.075000)
\psset{linewidth=0.100000cm}
\psset{linestyle=solid}
\psset{linestyle=solid}
\setlinejoinmode{0}
\newrgbcolor{dialinecolor}{0.000000 0.000000 0.000000}%
\psset{linecolor=dialinecolor}
\pspolygon(66.490400,10.075000)(66.490400,13.975000)(67.590400,13.975000)(67.590400,10.075000)
\setfont{Helvetica}{0.800000}
\newrgbcolor{dialinecolor}{0.000000 0.000000 0.000000}%
\psset{linecolor=dialinecolor}
\rput(67.040400,12.220000){\psscalebox{1 -1}{}}
\psset{linewidth=0.100000cm}
\psset{linestyle=solid}
\psset{linestyle=solid}
\setlinecaps{0}
\newrgbcolor{dialinecolor}{0.000000 0.000000 0.000000}%
\psset{linecolor=dialinecolor}
\psline(67.590400,13.975000)(71.685088,17.638830)
\psset{linestyle=solid}
\setlinejoinmode{0}
\setlinecaps{0}
\newrgbcolor{dialinecolor}{0.000000 0.000000 0.000000}%
\psset{linecolor=dialinecolor}
\pspolygon*(71.964548,17.888884)(71.425232,17.741786)(71.685088,17.638830)(71.758637,17.369172)
\newrgbcolor{dialinecolor}{0.000000 0.000000 0.000000}%
\psset{linecolor=dialinecolor}
\pspolygon(71.964548,17.888884)(71.425232,17.741786)(71.685088,17.638830)(71.758637,17.369172)
\setfont{Helvetica}{0.800000}
\newrgbcolor{dialinecolor}{0.000000 0.000000 0.000000}%
\psset{linecolor=dialinecolor}
\rput[l](24.596664,18.648282){\psscalebox{3 -3}{$h_1$}}
\setfont{Helvetica}{0.800000}
\newrgbcolor{dialinecolor}{0.000000 0.000000 0.000000}%
\psset{linecolor=dialinecolor}
\rput[l](18.656664,18.648282){\psscalebox{3 -3}{$h_0$}}
\setfont{Helvetica}{0.800000}
\newrgbcolor{dialinecolor}{0.000000 0.000000 0.000000}%
\psset{linecolor=dialinecolor}
\rput[l](30.673364,18.698282){\psscalebox{3 -3}{$h_2$}}
\setfont{Helvetica}{0.800000}
\newrgbcolor{dialinecolor}{0.000000 0.000000 0.000000}%
\psset{linecolor=dialinecolor}
\rput[l](36.663364,18.748282){\psscalebox{3 -3}{$h_3$}}
\setfont{Helvetica}{0.800000}
\newrgbcolor{dialinecolor}{0.000000 0.000000 0.000000}%
\psset{linecolor=dialinecolor}
\rput[l](51.573364,18.748282){\psscalebox{3 -3}{$h_{n-1}$}}
\setfont{Helvetica}{0.800000}
\newrgbcolor{dialinecolor}{0.000000 0.000000 0.000000}%
\psset{linecolor=dialinecolor}
\rput[l](57.893364,18.698282){\psscalebox{3 -3}{$h_n$}}
\setfont{Helvetica}{0.800000}
\newrgbcolor{dialinecolor}{0.000000 0.000000 0.000000}%
\psset{linecolor=dialinecolor}
\rput[l](72.723300,18.620000){\psscalebox{3 -3}{$h_1$}}
\setfont{Helvetica}{0.800000}
\newrgbcolor{dialinecolor}{0.000000 0.000000 0.000000}%
\psset{linecolor=dialinecolor}
\rput[l](78.773300,18.670000){\psscalebox{3 -3}{$h_2$}}
\setfont{Helvetica}{0.800000}
\newrgbcolor{dialinecolor}{0.000000 0.000000 0.000000}%
\psset{linecolor=dialinecolor}
\rput[l](84.780136,18.623300){\psscalebox{3 -3}{$h_3$}}
\setfont{Helvetica}{0.800000}
\newrgbcolor{dialinecolor}{0.000000 0.000000 0.000000}%
\psset{linecolor=dialinecolor}
\rput[l](97.390136,18.673300){\psscalebox{3 -3}{$h_{n-1}$}}
\setfont{Helvetica}{0.800000}
\newrgbcolor{dialinecolor}{0.000000 0.000000 0.000000}%
\psset{linecolor=dialinecolor}
\rput[l](103.710136,18.573300){\psscalebox{3 -3}{$h_n$}}
\setfont{Helvetica}{0.800000}
\newrgbcolor{dialinecolor}{0.000000 0.000000 0.000000}%
\psset{linecolor=dialinecolor}
\rput[l](24.603300,28.225000){\psscalebox{3 -3}{$x_1$}}
\setfont{Helvetica}{0.800000}
\newrgbcolor{dialinecolor}{0.000000 0.000000 0.000000}%
\psset{linecolor=dialinecolor}
\rput[l](30.703300,28.225000){\psscalebox{3 -3}{$x_2$}}
\setfont{Helvetica}{0.800000}
\newrgbcolor{dialinecolor}{0.000000 0.000000 0.000000}%
\psset{linecolor=dialinecolor}
\rput[l](36.653300,28.225000){\psscalebox{3 -3}{$x_3$}}
\setfont{Helvetica}{0.800000}
\newrgbcolor{dialinecolor}{0.000000 0.000000 0.000000}%
\psset{linecolor=dialinecolor}
\rput[l](51.603300,28.225000){\psscalebox{3 -3}{$x_{n-1}$}}
\setfont{Helvetica}{0.800000}
\newrgbcolor{dialinecolor}{0.000000 0.000000 0.000000}%
\psset{linecolor=dialinecolor}
\rput[l](57.903300,28.225000){\psscalebox{3 -3}{$x_n$}}
\setfont{Helvetica}{0.800000}
\newrgbcolor{dialinecolor}{0.000000 0.000000 0.000000}%
\psset{linecolor=dialinecolor}
\rput[l](72.703300,28.175000){\psscalebox{3 -3}{$x_1$}}
\setfont{Helvetica}{0.800000}
\newrgbcolor{dialinecolor}{0.000000 0.000000 0.000000}%
\psset{linecolor=dialinecolor}
\rput[l](78.753300,28.225000){\psscalebox{3 -3}{$x_2$}}
\setfont{Helvetica}{0.800000}
\newrgbcolor{dialinecolor}{0.000000 0.000000 0.000000}%
\psset{linecolor=dialinecolor}
\rput[l](84.853300,28.300000){\psscalebox{3 -3}{$x_3$}}
\setfont{Helvetica}{0.800000}
\newrgbcolor{dialinecolor}{0.000000 0.000000 0.000000}%
\psset{linecolor=dialinecolor}
\rput[l](97.453300,28.250000){\psscalebox{3 -3}{$x_{n-1}$}}
\setfont{Helvetica}{0.800000}
\newrgbcolor{dialinecolor}{0.000000 0.000000 0.000000}%
\psset{linecolor=dialinecolor}
\rput[l](103.713600,28.200000){\psscalebox{3 -3}{$x_n$}}
\setfont{Helvetica}{0.800000}
\newrgbcolor{dialinecolor}{0.000000 0.000000 0.000000}%
\psset{linecolor=dialinecolor}
\rput[l](24.603300,9.200000){\psscalebox{3 -3}{$o_1$}}
\setfont{Helvetica}{0.800000}
\newrgbcolor{dialinecolor}{0.000000 0.000000 0.000000}%
\psset{linecolor=dialinecolor}
\rput[l](30.603300,9.250000){\psscalebox{3 -3}{$o_2$}}
\setfont{Helvetica}{0.800000}
\newrgbcolor{dialinecolor}{0.000000 0.000000 0.000000}%
\psset{linecolor=dialinecolor}
\rput[l](36.603300,9.250000){\psscalebox{3 -3}{$o_3$}}
\setfont{Helvetica}{0.800000}
\newrgbcolor{dialinecolor}{0.000000 0.000000 0.000000}%
\psset{linecolor=dialinecolor}
\rput[l](51.503300,9.250000){\psscalebox{3 -3}{$o_{n-1}$}}
\setfont{Helvetica}{0.800000}
\newrgbcolor{dialinecolor}{0.000000 0.000000 0.000000}%
\psset{linecolor=dialinecolor}
\rput[l](57.903300,9.200000){\psscalebox{3 -3}{$o_n$}}
\setfont{Helvetica}{0.800000}
\newrgbcolor{dialinecolor}{0.000000 0.000000 0.000000}%
\psset{linecolor=dialinecolor}
\rput[l](66.703300,9.250000){\psscalebox{3 -3}{$o_0$}}
\setfont{Helvetica}{0.800000}
\newrgbcolor{dialinecolor}{0.000000 0.000000 0.000000}%
\psset{linecolor=dialinecolor}
\rput[l](72.703300,9.300000){\psscalebox{3 -3}{$o_1$}}
\setfont{Helvetica}{0.800000}
\newrgbcolor{dialinecolor}{0.000000 0.000000 0.000000}%
\psset{linecolor=dialinecolor}
\rput[l](78.753300,9.300000){\psscalebox{3 -3}{$o_2$}}
\setfont{Helvetica}{0.800000}
\newrgbcolor{dialinecolor}{0.000000 0.000000 0.000000}%
\psset{linecolor=dialinecolor}
\rput[l](84.853300,9.250000){\psscalebox{3 -3}{$o_3$}}
\setfont{Helvetica}{0.800000}
\newrgbcolor{dialinecolor}{0.000000 0.000000 0.000000}%
\psset{linecolor=dialinecolor}
\rput[l](97.453300,9.250000){\psscalebox{3 -3}{$o_{n-1}$}}
\setfont{Helvetica}{0.800000}
\newrgbcolor{dialinecolor}{0.000000 0.000000 0.000000}%
\psset{linecolor=dialinecolor}
\rput[l](103.613600,9.150000){\psscalebox{3 -3}{$o_n$}}
\setfont{Helvetica}{0.800000}
\newrgbcolor{dialinecolor}{0.000000 0.000000 0.000000}%
\psset{linecolor=dialinecolor}
\rput[l](37.453300,30.600000){\psscalebox{3 -3}{ELMAN}}
\setfont{Helvetica}{0.800000}
\newrgbcolor{dialinecolor}{0.000000 0.000000 0.000000}%
\psset{linecolor=dialinecolor}
\rput[l](88.253300,30.550000){\psscalebox{3 -3}{JORDAN}}
}\endpspicture

%% file: wordEmbeddings.tex
\ifx\setlinejoinmode\undefined
  \newcommand{\setlinejoinmode}[1]{}
\fi
\ifx\setlinecaps\undefined
  \newcommand{\setlinecaps}[1]{}
\fi
\ifx\setfont\undefined
  \newcommand{\setfont}[2]{}
\fi
\pspicture(0.786501,-7.493934)(16.142101,-1.408178)
\psscalebox{0.185059 -0.185059}{
\newrgbcolor{dialinecolor}{0.000000 0.000000 0.000000}%
\psset{linecolor=dialinecolor}
\newrgbcolor{diafillcolor}{1.000000 1.000000 1.000000}%
\psset{fillcolor=diafillcolor}
\newrgbcolor{dialinecolor}{1.000000 1.000000 1.000000}%
\psset{linecolor=dialinecolor}
\pspolygon*(8.100000,9.800000)(8.100000,13.750000)(9.950000,13.750000)(9.950000,9.800000)
\psset{linewidth=0.100000cm}
\psset{linestyle=solid}
\psset{linestyle=solid}
\setlinejoinmode{0}
\newrgbcolor{dialinecolor}{0.000000 0.000000 0.000000}%
\psset{linecolor=dialinecolor}
\pspolygon(8.100000,9.800000)(8.100000,13.750000)(9.950000,13.750000)(9.950000,9.800000)
\setfont{Helvetica}{0.800000}
\newrgbcolor{dialinecolor}{0.000000 0.000000 0.000000}%
\psset{linecolor=dialinecolor}
\rput(9.025000,11.970000){\psscalebox{1 -1}{}}
\newrgbcolor{dialinecolor}{1.000000 1.000000 1.000000}%
\psset{linecolor=dialinecolor}
\pspolygon*(8.120000,16.050000)(8.120000,20.000000)(9.970000,20.000000)(9.970000,16.050000)
\psset{linewidth=0.100000cm}
\psset{linestyle=solid}
\psset{linestyle=solid}
\setlinejoinmode{0}
\newrgbcolor{dialinecolor}{0.000000 0.000000 0.000000}%
\psset{linecolor=dialinecolor}
\pspolygon(8.120000,16.050000)(8.120000,20.000000)(9.970000,20.000000)(9.970000,16.050000)
\setfont{Helvetica}{0.800000}
\newrgbcolor{dialinecolor}{0.000000 0.000000 0.000000}%
\psset{linecolor=dialinecolor}
\rput(9.045000,18.220000){\psscalebox{1 -1}{}}
\newrgbcolor{dialinecolor}{1.000000 1.000000 1.000000}%
\psset{linecolor=dialinecolor}
\pspolygon*(8.070000,26.150000)(8.070000,30.100000)(9.920000,30.100000)(9.920000,26.150000)
\psset{linewidth=0.100000cm}
\psset{linestyle=solid}
\psset{linestyle=solid}
\setlinejoinmode{0}
\newrgbcolor{dialinecolor}{0.000000 0.000000 0.000000}%
\psset{linecolor=dialinecolor}
\pspolygon(8.070000,26.150000)(8.070000,30.100000)(9.920000,30.100000)(9.920000,26.150000)
\setfont{Helvetica}{0.800000}
\newrgbcolor{dialinecolor}{0.000000 0.000000 0.000000}%
\psset{linecolor=dialinecolor}
\rput(8.995000,28.320000){\psscalebox{1 -1}{}}
\newrgbcolor{dialinecolor}{1.000000 1.000000 1.000000}%
\psset{linecolor=dialinecolor}
\pspolygon*(8.090000,32.400000)(8.090000,36.350000)(9.940000,36.350000)(9.940000,32.400000)
\psset{linewidth=0.100000cm}
\psset{linestyle=solid}
\psset{linestyle=solid}
\setlinejoinmode{0}
\newrgbcolor{dialinecolor}{0.000000 0.000000 0.000000}%
\psset{linecolor=dialinecolor}
\pspolygon(8.090000,32.400000)(8.090000,36.350000)(9.940000,36.350000)(9.940000,32.400000)
\setfont{Helvetica}{0.800000}
\newrgbcolor{dialinecolor}{0.000000 0.000000 0.000000}%
\psset{linecolor=dialinecolor}
\rput(9.015000,34.570000){\psscalebox{1 -1}{}}
\newrgbcolor{dialinecolor}{0.662745 0.662745 0.662745}%
\psset{linecolor=dialinecolor}
\pspolygon*(16.120000,21.000000)(16.120000,24.950000)(17.970000,24.950000)(17.970000,21.000000)
\psset{linewidth=0.100000cm}
\psset{linestyle=solid}
\psset{linestyle=solid}
\setlinejoinmode{0}
\newrgbcolor{dialinecolor}{0.000000 0.000000 0.000000}%
\psset{linecolor=dialinecolor}
\pspolygon(16.120000,21.000000)(16.120000,24.950000)(17.970000,24.950000)(17.970000,21.000000)
\setfont{Helvetica}{0.800000}
\newrgbcolor{dialinecolor}{0.000000 0.000000 0.000000}%
\psset{linecolor=dialinecolor}
\rput(17.045000,23.170000){\psscalebox{1 -1}{}}
\newrgbcolor{dialinecolor}{1.000000 1.000000 1.000000}%
\psset{linecolor=dialinecolor}
\pspolygon*(24.140000,21.029300)(24.140000,24.979300)(25.990000,24.979300)(25.990000,21.029300)
\psset{linewidth=0.100000cm}
\psset{linestyle=solid}
\psset{linestyle=solid}
\setlinejoinmode{0}
\newrgbcolor{dialinecolor}{0.000000 0.000000 0.000000}%
\psset{linecolor=dialinecolor}
\pspolygon(24.140000,21.029300)(24.140000,24.979300)(25.990000,24.979300)(25.990000,21.029300)
\setfont{Helvetica}{0.800000}
\newrgbcolor{dialinecolor}{0.000000 0.000000 0.000000}%
\psset{linecolor=dialinecolor}
\rput(25.065000,23.199300){\psscalebox{1 -1}{}}
\psset{linewidth=0.100000cm}
\psset{linestyle=solid}
\psset{linestyle=solid}
\setlinecaps{0}
\newrgbcolor{dialinecolor}{0.000000 0.000000 0.000000}%
\psset{linecolor=dialinecolor}
\psline(9.940000,34.375000)(15.853068,25.357093)
\psset{linestyle=solid}
\setlinejoinmode{0}
\setlinecaps{0}
\newrgbcolor{dialinecolor}{0.000000 0.000000 0.000000}%
\psset{linecolor=dialinecolor}
\pspolygon*(16.058694,25.043496)(15.993590,25.598709)(15.853068,25.357093)(15.575461,25.324541)
\newrgbcolor{dialinecolor}{0.000000 0.000000 0.000000}%
\psset{linecolor=dialinecolor}
\pspolygon(16.058694,25.043496)(15.993590,25.598709)(15.853068,25.357093)(15.575461,25.324541)
\psset{linewidth=0.100000cm}
\psset{linestyle=solid}
\psset{linestyle=solid}
\setlinecaps{0}
\newrgbcolor{dialinecolor}{0.000000 0.000000 0.000000}%
\psset{linecolor=dialinecolor}
\psline(9.920000,28.125000)(15.715835,24.233845)
\psset{linestyle=solid}
\setlinejoinmode{0}
\setlinecaps{0}
\newrgbcolor{dialinecolor}{0.000000 0.000000 0.000000}%
\psset{linecolor=dialinecolor}
\pspolygon*(16.027176,24.024819)(15.751405,24.511081)(15.715835,24.233845)(15.472704,24.095959)
\newrgbcolor{dialinecolor}{0.000000 0.000000 0.000000}%
\psset{linecolor=dialinecolor}
\pspolygon(16.027176,24.024819)(15.751405,24.511081)(15.715835,24.233845)(15.472704,24.095959)
\psset{linewidth=0.100000cm}
\psset{linestyle=solid}
\psset{linestyle=solid}
\setlinecaps{0}
\newrgbcolor{dialinecolor}{0.000000 0.000000 0.000000}%
\psset{linecolor=dialinecolor}
\psline(9.970000,18.025000)(15.710782,21.723837)
\psset{linestyle=solid}
\setlinejoinmode{0}
\setlinecaps{0}
\newrgbcolor{dialinecolor}{0.000000 0.000000 0.000000}%
\psset{linecolor=dialinecolor}
\pspolygon*(16.026016,21.926945)(15.470299,21.866290)(15.710782,21.723837)(15.741109,21.445979)
\newrgbcolor{dialinecolor}{0.000000 0.000000 0.000000}%
\psset{linecolor=dialinecolor}
\pspolygon(16.026016,21.926945)(15.470299,21.866290)(15.710782,21.723837)(15.741109,21.445979)
\psset{linewidth=0.100000cm}
\psset{linestyle=solid}
\psset{linestyle=solid}
\setlinecaps{0}
\newrgbcolor{dialinecolor}{0.000000 0.000000 0.000000}%
\psset{linecolor=dialinecolor}
\psline(9.950000,11.775000)(15.849363,20.595360)
\psset{linestyle=solid}
\setlinejoinmode{0}
\setlinecaps{0}
\newrgbcolor{dialinecolor}{0.000000 0.000000 0.000000}%
\psset{linecolor=dialinecolor}
\pspolygon*(16.057843,20.907067)(15.572065,20.630445)(15.849363,20.595360)(15.987674,20.352471)
\newrgbcolor{dialinecolor}{0.000000 0.000000 0.000000}%
\psset{linecolor=dialinecolor}
\pspolygon(16.057843,20.907067)(15.572065,20.630445)(15.849363,20.595360)(15.987674,20.352471)
\psset{linewidth=0.100000cm}
\psset{linestyle=solid}
\psset{linestyle=solid}
\setlinecaps{0}
\newrgbcolor{dialinecolor}{0.000000 0.000000 0.000000}%
\psset{linecolor=dialinecolor}
\psline(17.970000,22.975000)(23.653202,23.001988)
\psset{linestyle=solid}
\setlinejoinmode{0}
\setlinecaps{0}
\newrgbcolor{dialinecolor}{0.000000 0.000000 0.000000}%
\psset{linecolor=dialinecolor}
\pspolygon*(24.028198,23.003769)(23.527016,23.251392)(23.653202,23.001988)(23.529391,22.751398)
\newrgbcolor{dialinecolor}{0.000000 0.000000 0.000000}%
\psset{linecolor=dialinecolor}
\pspolygon(24.028198,23.003769)(23.527016,23.251392)(23.653202,23.001988)(23.529391,22.751398)
\setfont{Helvetica}{0.800000}
\newrgbcolor{dialinecolor}{0.000000 0.000000 0.000000}%
\psset{linecolor=dialinecolor}
\rput[l](4.600000,11.700000){\psscalebox{4 -4}{$w_{t-2}$}}
\setfont{Helvetica}{0.800000}
\newrgbcolor{dialinecolor}{0.000000 0.000000 0.000000}%
\psset{linecolor=dialinecolor}
\rput[l](4.650000,17.800000){\psscalebox{4 -4}{$w_{t-1}$}}
\setfont{Helvetica}{0.800000}
\newrgbcolor{dialinecolor}{0.000000 0.000000 0.000000}%
\psset{linecolor=dialinecolor}
\rput[l](4.250000,27.950000){\psscalebox{4 -4}{$w_{t+1}$}}
\setfont{Helvetica}{0.800000}
\newrgbcolor{dialinecolor}{0.000000 0.000000 0.000000}%
\psset{linecolor=dialinecolor}
\rput[l](4.350000,34.200000){\psscalebox{4 -4}{$w_{t+2}$}}
\setfont{Helvetica}{0.800000}
\newrgbcolor{dialinecolor}{0.000000 0.000000 0.000000}%
\psset{linecolor=dialinecolor}
\rput[l](26.400000,22.950000){\psscalebox{4 -4}{$w_t$}}
\setfont{Helvetica}{0.800000}
\newrgbcolor{dialinecolor}{0.000000 0.000000 0.000000}%
\psset{linecolor=dialinecolor}
\rput[l](16.150000,19.551500){\psscalebox{3 -3}{SUM}}
\setfont{Helvetica}{0.800000}
\newrgbcolor{dialinecolor}{0.000000 0.000000 0.000000}%
\psset{linecolor=dialinecolor}
\rput[l](7.300000,8.300000){\psscalebox{3 -3}{INPUT}}
\setfont{Helvetica}{0.800000}
\newrgbcolor{dialinecolor}{0.000000 0.000000 0.000000}%
\psset{linecolor=dialinecolor}
\rput[l](14.050000,8.250000){\psscalebox{3 -3}{PROJECTION}}
\setfont{Helvetica}{0.800000}
\newrgbcolor{dialinecolor}{0.000000 0.000000 0.000000}%
\psset{linecolor=dialinecolor}
\rput[l](22.950000,8.300000){\psscalebox{3 -3}{OUTPUT}}
\newrgbcolor{dialinecolor}{1.000000 1.000000 1.000000}%
\psset{linecolor=dialinecolor}
\pspolygon*(52.529000,9.795000)(52.529000,13.745000)(54.379000,13.745000)(54.379000,9.795000)
\psset{linewidth=0.100000cm}
\psset{linestyle=solid}
\psset{linestyle=solid}
\setlinejoinmode{0}
\newrgbcolor{dialinecolor}{0.000000 0.000000 0.000000}%
\psset{linecolor=dialinecolor}
\pspolygon(52.529000,9.795000)(52.529000,13.745000)(54.379000,13.745000)(54.379000,9.795000)
\setfont{Helvetica}{0.800000}
\newrgbcolor{dialinecolor}{0.000000 0.000000 0.000000}%
\psset{linecolor=dialinecolor}
\rput(53.454000,11.965000){\psscalebox{1 -1}{}}
\newrgbcolor{dialinecolor}{1.000000 1.000000 1.000000}%
\psset{linecolor=dialinecolor}
\pspolygon*(52.549000,16.045000)(52.549000,19.995000)(54.399000,19.995000)(54.399000,16.045000)
\psset{linewidth=0.100000cm}
\psset{linestyle=solid}
\psset{linestyle=solid}
\setlinejoinmode{0}
\newrgbcolor{dialinecolor}{0.000000 0.000000 0.000000}%
\psset{linecolor=dialinecolor}
\pspolygon(52.549000,16.045000)(52.549000,19.995000)(54.399000,19.995000)(54.399000,16.045000)
\setfont{Helvetica}{0.800000}
\newrgbcolor{dialinecolor}{0.000000 0.000000 0.000000}%
\psset{linecolor=dialinecolor}
\rput(53.474000,18.215000){\psscalebox{1 -1}{}}
\newrgbcolor{dialinecolor}{1.000000 1.000000 1.000000}%
\psset{linecolor=dialinecolor}
\pspolygon*(52.499000,26.145000)(52.499000,30.095000)(54.349000,30.095000)(54.349000,26.145000)
\psset{linewidth=0.100000cm}
\psset{linestyle=solid}
\psset{linestyle=solid}
\setlinejoinmode{0}
\newrgbcolor{dialinecolor}{0.000000 0.000000 0.000000}%
\psset{linecolor=dialinecolor}
\pspolygon(52.499000,26.145000)(52.499000,30.095000)(54.349000,30.095000)(54.349000,26.145000)
\setfont{Helvetica}{0.800000}
\newrgbcolor{dialinecolor}{0.000000 0.000000 0.000000}%
\psset{linecolor=dialinecolor}
\rput(53.424000,28.315000){\psscalebox{1 -1}{}}
\newrgbcolor{dialinecolor}{1.000000 1.000000 1.000000}%
\psset{linecolor=dialinecolor}
\pspolygon*(52.519000,32.395000)(52.519000,36.345000)(54.369000,36.345000)(54.369000,32.395000)
\psset{linewidth=0.100000cm}
\psset{linestyle=solid}
\psset{linestyle=solid}
\setlinejoinmode{0}
\newrgbcolor{dialinecolor}{0.000000 0.000000 0.000000}%
\psset{linecolor=dialinecolor}
\pspolygon(52.519000,32.395000)(52.519000,36.345000)(54.369000,36.345000)(54.369000,32.395000)
\setfont{Helvetica}{0.800000}
\newrgbcolor{dialinecolor}{0.000000 0.000000 0.000000}%
\psset{linecolor=dialinecolor}
\rput(53.444000,34.565000){\psscalebox{1 -1}{}}
\newrgbcolor{dialinecolor}{0.662745 0.662745 0.662745}%
\psset{linecolor=dialinecolor}
\pspolygon*(44.497700,20.995000)(44.497700,24.945000)(46.347700,24.945000)(46.347700,20.995000)
\psset{linewidth=0.100000cm}
\psset{linestyle=solid}
\psset{linestyle=solid}
\setlinejoinmode{0}
\newrgbcolor{dialinecolor}{0.000000 0.000000 0.000000}%
\psset{linecolor=dialinecolor}
\pspolygon(44.497700,20.995000)(44.497700,24.945000)(46.347700,24.945000)(46.347700,20.995000)
\setfont{Helvetica}{0.800000}
\newrgbcolor{dialinecolor}{0.000000 0.000000 0.000000}%
\psset{linecolor=dialinecolor}
\rput(45.422700,23.165000){\psscalebox{1 -1}{}}
\newrgbcolor{dialinecolor}{1.000000 1.000000 1.000000}%
\psset{linecolor=dialinecolor}
\pspolygon*(36.183500,21.024300)(36.183500,24.974300)(38.033500,24.974300)(38.033500,21.024300)
\psset{linewidth=0.100000cm}
\psset{linestyle=solid}
\psset{linestyle=solid}
\setlinejoinmode{0}
\newrgbcolor{dialinecolor}{0.000000 0.000000 0.000000}%
\psset{linecolor=dialinecolor}
\pspolygon(36.183500,21.024300)(36.183500,24.974300)(38.033500,24.974300)(38.033500,21.024300)
\setfont{Helvetica}{0.800000}
\newrgbcolor{dialinecolor}{0.000000 0.000000 0.000000}%
\psset{linecolor=dialinecolor}
\rput(37.108500,23.194300){\psscalebox{1 -1}{}}
\setfont{Helvetica}{0.800000}
\newrgbcolor{dialinecolor}{0.000000 0.000000 0.000000}%
\psset{linecolor=dialinecolor}
\rput[l](54.870200,11.695000){\psscalebox{4 -4}{$w_{t-2}$}}
\setfont{Helvetica}{0.800000}
\newrgbcolor{dialinecolor}{0.000000 0.000000 0.000000}%
\psset{linecolor=dialinecolor}
\rput[l](54.920200,17.795000){\psscalebox{4 -4}{$w_{t-1}$}}
\setfont{Helvetica}{0.800000}
\newrgbcolor{dialinecolor}{0.000000 0.000000 0.000000}%
\psset{linecolor=dialinecolor}
\rput[l](54.732300,27.945000){\psscalebox{4 -4}{$w_{t+1}$}}
\setfont{Helvetica}{0.800000}
\newrgbcolor{dialinecolor}{0.000000 0.000000 0.000000}%
\psset{linecolor=dialinecolor}
\rput[l](54.832300,34.195000){\psscalebox{4 -4}{$w_{t+2}$}}
\setfont{Helvetica}{0.800000}
\newrgbcolor{dialinecolor}{0.000000 0.000000 0.000000}%
\psset{linecolor=dialinecolor}
\rput[l](34.266600,22.945000){\psscalebox{4 -4}{$w_t$}}
\setfont{Helvetica}{0.800000}
\newrgbcolor{dialinecolor}{0.000000 0.000000 0.000000}%
\psset{linecolor=dialinecolor}
\rput[l](35.515600,8.295000){\psscalebox{3 -3}{INPUT}}
\setfont{Helvetica}{0.800000}
\newrgbcolor{dialinecolor}{0.000000 0.000000 0.000000}%
\psset{linecolor=dialinecolor}
\rput[l](42.477700,8.245000){\psscalebox{3 -3}{PROJECTION}}
\setfont{Helvetica}{0.800000}
\newrgbcolor{dialinecolor}{0.000000 0.000000 0.000000}%
\psset{linecolor=dialinecolor}
\rput[l](51.377700,8.295000){\psscalebox{3 -3}{OUTPUT}}
\psset{linewidth=0.100000cm}
\psset{linestyle=solid}
\psset{linestyle=solid}
\setlinecaps{0}
\newrgbcolor{dialinecolor}{0.000000 0.000000 0.000000}%
\psset{linecolor=dialinecolor}
\psline(46.347700,20.995000)(52.258021,12.174410)
\psset{linestyle=solid}
\setlinejoinmode{0}
\setlinecaps{0}
\newrgbcolor{dialinecolor}{0.000000 0.000000 0.000000}%
\psset{linecolor=dialinecolor}
\pspolygon*(52.466765,11.862880)(52.396126,12.417416)(52.258021,12.174410)(51.980753,12.139091)
\newrgbcolor{dialinecolor}{0.000000 0.000000 0.000000}%
\psset{linecolor=dialinecolor}
\pspolygon(52.466765,11.862880)(52.396126,12.417416)(52.258021,12.174410)(51.980753,12.139091)
\psset{linewidth=0.100000cm}
\psset{linestyle=solid}
\psset{linestyle=solid}
\setlinecaps{0}
\newrgbcolor{dialinecolor}{0.000000 0.000000 0.000000}%
\psset{linecolor=dialinecolor}
\psline(46.347700,21.982500)(52.138789,18.282116)
\psset{linestyle=solid}
\setlinejoinmode{0}
\setlinecaps{0}
\newrgbcolor{dialinecolor}{0.000000 0.000000 0.000000}%
\psset{linecolor=dialinecolor}
\pspolygon*(52.454788,18.080200)(52.168067,18.560087)(52.138789,18.282116)(51.898846,18.138756)
\newrgbcolor{dialinecolor}{0.000000 0.000000 0.000000}%
\psset{linecolor=dialinecolor}
\pspolygon(52.454788,18.080200)(52.168067,18.560087)(52.138789,18.282116)(51.898846,18.138756)
\psset{linewidth=0.100000cm}
\psset{linestyle=solid}
\psset{linestyle=solid}
\setlinecaps{0}
\newrgbcolor{dialinecolor}{0.000000 0.000000 0.000000}%
\psset{linecolor=dialinecolor}
\psline(46.347700,23.957500)(52.095829,27.847180)
\psset{linestyle=solid}
\setlinejoinmode{0}
\setlinecaps{0}
\newrgbcolor{dialinecolor}{0.000000 0.000000 0.000000}%
\psset{linecolor=dialinecolor}
\pspolygon*(52.406404,28.057342)(51.852196,27.984176)(52.095829,27.847180)(52.132412,27.570076)
\newrgbcolor{dialinecolor}{0.000000 0.000000 0.000000}%
\psset{linecolor=dialinecolor}
\pspolygon(52.406404,28.057342)(51.852196,27.984176)(52.095829,27.847180)(52.132412,27.570076)
\psset{linewidth=0.100000cm}
\psset{linestyle=solid}
\psset{linestyle=solid}
\setlinecaps{0}
\newrgbcolor{dialinecolor}{0.000000 0.000000 0.000000}%
\psset{linecolor=dialinecolor}
\psline(46.347700,24.945000)(52.252331,33.962735)
\psset{linestyle=solid}
\setlinejoinmode{0}
\setlinecaps{0}
\newrgbcolor{dialinecolor}{0.000000 0.000000 0.000000}%
\psset{linecolor=dialinecolor}
\pspolygon*(52.457754,34.276464)(51.974703,33.995107)(52.252331,33.962735)(52.393009,33.721209)
\newrgbcolor{dialinecolor}{0.000000 0.000000 0.000000}%
\psset{linecolor=dialinecolor}
\pspolygon(52.457754,34.276464)(51.974703,33.995107)(52.252331,33.962735)(52.393009,33.721209)
\psset{linewidth=0.100000cm}
\psset{linestyle=solid}
\psset{linestyle=solid}
\setlinecaps{0}
\newrgbcolor{dialinecolor}{0.000000 0.000000 0.000000}%
\psset{linecolor=dialinecolor}
\psline(38.033500,22.999300)(44.010902,22.972206)
\psset{linestyle=solid}
\setlinejoinmode{0}
\setlinecaps{0}
\newrgbcolor{dialinecolor}{0.000000 0.000000 0.000000}%
\psset{linecolor=dialinecolor}
\pspolygon*(44.385898,22.970507)(43.887036,23.222770)(44.010902,22.972206)(43.884770,22.722776)
\newrgbcolor{dialinecolor}{0.000000 0.000000 0.000000}%
\psset{linecolor=dialinecolor}
\pspolygon(44.385898,22.970507)(43.887036,23.222770)(44.010902,22.972206)(43.884770,22.722776)
\newrgbcolor{dialinecolor}{1.000000 1.000000 1.000000}%
\psset{linecolor=dialinecolor}
\pspolygon*(67.884200,9.754340)(67.884200,13.704340)(69.734200,13.704340)(69.734200,9.754340)
\psset{linewidth=0.100000cm}
\psset{linestyle=solid}
\psset{linestyle=solid}
\setlinejoinmode{0}
\newrgbcolor{dialinecolor}{0.000000 0.000000 0.000000}%
\psset{linecolor=dialinecolor}
\pspolygon(67.884200,9.754340)(67.884200,13.704340)(69.734200,13.704340)(69.734200,9.754340)
\setfont{Helvetica}{0.800000}
\newrgbcolor{dialinecolor}{0.000000 0.000000 0.000000}%
\psset{linecolor=dialinecolor}
\rput(68.809200,11.924340){\psscalebox{1 -1}{}}
\newrgbcolor{dialinecolor}{1.000000 1.000000 1.000000}%
\psset{linecolor=dialinecolor}
\pspolygon*(67.904200,16.004300)(67.904200,19.954300)(69.754200,19.954300)(69.754200,16.004300)
\psset{linewidth=0.100000cm}
\psset{linestyle=solid}
\psset{linestyle=solid}
\setlinejoinmode{0}
\newrgbcolor{dialinecolor}{0.000000 0.000000 0.000000}%
\psset{linecolor=dialinecolor}
\pspolygon(67.904200,16.004300)(67.904200,19.954300)(69.754200,19.954300)(69.754200,16.004300)
\setfont{Helvetica}{0.800000}
\newrgbcolor{dialinecolor}{0.000000 0.000000 0.000000}%
\psset{linecolor=dialinecolor}
\rput(68.829200,18.174300){\psscalebox{1 -1}{}}
\newrgbcolor{dialinecolor}{1.000000 1.000000 1.000000}%
\psset{linecolor=dialinecolor}
\pspolygon*(67.854200,26.104300)(67.854200,30.054300)(69.704200,30.054300)(69.704200,26.104300)
\psset{linewidth=0.100000cm}
\psset{linestyle=solid}
\psset{linestyle=solid}
\setlinejoinmode{0}
\newrgbcolor{dialinecolor}{0.000000 0.000000 0.000000}%
\psset{linecolor=dialinecolor}
\pspolygon(67.854200,26.104300)(67.854200,30.054300)(69.704200,30.054300)(69.704200,26.104300)
\setfont{Helvetica}{0.800000}
\newrgbcolor{dialinecolor}{0.000000 0.000000 0.000000}%
\psset{linecolor=dialinecolor}
\rput(68.779200,28.274300){\psscalebox{1 -1}{}}
\newrgbcolor{dialinecolor}{1.000000 1.000000 1.000000}%
\psset{linecolor=dialinecolor}
\pspolygon*(67.874200,32.354300)(67.874200,36.304300)(69.724200,36.304300)(69.724200,32.354300)
\psset{linewidth=0.100000cm}
\psset{linestyle=solid}
\psset{linestyle=solid}
\setlinejoinmode{0}
\newrgbcolor{dialinecolor}{0.000000 0.000000 0.000000}%
\psset{linecolor=dialinecolor}
\pspolygon(67.874200,32.354300)(67.874200,36.304300)(69.724200,36.304300)(69.724200,32.354300)
\setfont{Helvetica}{0.800000}
\newrgbcolor{dialinecolor}{0.000000 0.000000 0.000000}%
\psset{linecolor=dialinecolor}
\rput(68.799200,34.524300){\psscalebox{1 -1}{}}
\newrgbcolor{dialinecolor}{1.000000 1.000000 1.000000}%
\psset{linecolor=dialinecolor}
\pspolygon*(83.924200,21.054300)(83.924200,25.004300)(85.774200,25.004300)(85.774200,21.054300)
\psset{linewidth=0.100000cm}
\psset{linestyle=solid}
\psset{linestyle=solid}
\setlinejoinmode{0}
\newrgbcolor{dialinecolor}{0.000000 0.000000 0.000000}%
\psset{linecolor=dialinecolor}
\pspolygon(83.924200,21.054300)(83.924200,25.004300)(85.774200,25.004300)(85.774200,21.054300)
\setfont{Helvetica}{0.800000}
\newrgbcolor{dialinecolor}{0.000000 0.000000 0.000000}%
\psset{linecolor=dialinecolor}
\rput(84.849200,23.224300){\psscalebox{1 -1}{}}
\setfont{Helvetica}{0.800000}
\newrgbcolor{dialinecolor}{0.000000 0.000000 0.000000}%
\psset{linecolor=dialinecolor}
\rput[l](64.234200,11.654300){\psscalebox{4 -4}{$w_{t-2}$}}
\setfont{Helvetica}{0.800000}
\newrgbcolor{dialinecolor}{0.000000 0.000000 0.000000}%
\psset{linecolor=dialinecolor}
\rput[l](64.284200,17.754300){\psscalebox{4 -4}{$w_{t-1}$}}
\setfont{Helvetica}{0.800000}
\newrgbcolor{dialinecolor}{0.000000 0.000000 0.000000}%
\psset{linecolor=dialinecolor}
\rput[l](63.884200,27.904300){\psscalebox{4 -4}{$w_{t+1}$}}
\setfont{Helvetica}{0.800000}
\newrgbcolor{dialinecolor}{0.000000 0.000000 0.000000}%
\psset{linecolor=dialinecolor}
\rput[l](63.984200,34.154300){\psscalebox{4 -4}{$w_{t+2}$}}
\setfont{Helvetica}{0.800000}
\newrgbcolor{dialinecolor}{0.000000 0.000000 0.000000}%
\psset{linecolor=dialinecolor}
\rput[l](86.134200,22.904300){\psscalebox{4 -4}{$w_t$}}
\setfont{Helvetica}{0.800000}
\newrgbcolor{dialinecolor}{0.000000 0.000000 0.000000}%
\psset{linecolor=dialinecolor}
\rput[l](67.284200,8.254340){\psscalebox{3 -3}{INPUT}}
\setfont{Helvetica}{0.800000}
\newrgbcolor{dialinecolor}{0.000000 0.000000 0.000000}%
\psset{linecolor=dialinecolor}
\rput[l](73.734200,8.204340){\psscalebox{3 -3}{PROJECTION}}
\setfont{Helvetica}{0.800000}
\newrgbcolor{dialinecolor}{0.000000 0.000000 0.000000}%
\psset{linecolor=dialinecolor}
\rput[l](82.784200,8.254340){\psscalebox{3 -3}{OUTPUT}}
\newrgbcolor{dialinecolor}{0.662745 0.662745 0.662745}%
\psset{linecolor=dialinecolor}
\pspolygon*(75.928200,15.168900)(75.928200,30.937381)(77.766678,30.937381)(77.766678,15.168900)
\psset{linewidth=0.100000cm}
\psset{linestyle=solid}
\psset{linestyle=solid}
\setlinejoinmode{0}
\newrgbcolor{dialinecolor}{0.000000 0.000000 0.000000}%
\psset{linecolor=dialinecolor}
\pspolygon(75.928200,15.168900)(75.928200,30.937381)(77.766678,30.937381)(77.766678,15.168900)
\setfont{Helvetica}{0.800000}
\newrgbcolor{dialinecolor}{0.000000 0.000000 0.000000}%
\psset{linecolor=dialinecolor}
\rput(76.847439,23.248141){\psscalebox{1 -1}{}}
\psset{linewidth=0.100000cm}
\psset{linestyle=solid}
\psset{linestyle=solid}
\setlinecaps{0}
\newrgbcolor{dialinecolor}{0.000000 0.000000 0.000000}%
\psset{linecolor=dialinecolor}
\psline(77.766700,23.053100)(83.437400,23.031182)
\psset{linestyle=solid}
\setlinejoinmode{0}
\setlinecaps{0}
\newrgbcolor{dialinecolor}{0.000000 0.000000 0.000000}%
\psset{linecolor=dialinecolor}
\pspolygon*(83.812397,23.029732)(83.313367,23.281663)(83.437400,23.031182)(83.311435,22.781667)
\newrgbcolor{dialinecolor}{0.000000 0.000000 0.000000}%
\psset{linecolor=dialinecolor}
\pspolygon(83.812397,23.029732)(83.313367,23.281663)(83.437400,23.031182)(83.311435,22.781667)
\psset{linewidth=0.100000cm}
\psset{linestyle=solid}
\psset{linestyle=solid}
\setlinecaps{0}
\newrgbcolor{dialinecolor}{0.000000 0.000000 0.000000}%
\psset{linecolor=dialinecolor}
\psline(69.734200,11.729300)(75.502613,14.932567)
\psset{linestyle=solid}
\setlinejoinmode{0}
\setlinecaps{0}
\newrgbcolor{dialinecolor}{0.000000 0.000000 0.000000}%
\psset{linecolor=dialinecolor}
\pspolygon*(75.830456,15.114622)(75.271962,15.090444)(75.502613,14.932567)(75.514702,14.653320)
\newrgbcolor{dialinecolor}{0.000000 0.000000 0.000000}%
\psset{linecolor=dialinecolor}
\pspolygon(75.830456,15.114622)(75.271962,15.090444)(75.502613,14.932567)(75.514702,14.653320)
\psset{linewidth=0.100000cm}
\psset{linestyle=solid}
\psset{linestyle=solid}
\setlinecaps{0}
\newrgbcolor{dialinecolor}{0.000000 0.000000 0.000000}%
\psset{linecolor=dialinecolor}
\psline(69.724200,34.329300)(75.501066,31.170926)
\psset{linestyle=solid}
\setlinejoinmode{0}
\setlinecaps{0}
\newrgbcolor{dialinecolor}{0.000000 0.000000 0.000000}%
\psset{linecolor=dialinecolor}
\pspolygon*(75.830101,30.991034)(75.511316,31.450246)(75.501066,31.170926)(75.271460,31.011534)
\newrgbcolor{dialinecolor}{0.000000 0.000000 0.000000}%
\psset{linecolor=dialinecolor}
\pspolygon(75.830101,30.991034)(75.511316,31.450246)(75.501066,31.170926)(75.271460,31.011534)
\psset{linewidth=0.100000cm}
\psset{linestyle=solid}
\psset{linestyle=solid}
\setlinecaps{0}
\newrgbcolor{dialinecolor}{0.000000 0.000000 0.000000}%
\psset{linecolor=dialinecolor}
\psline(69.754200,17.979300)(75.449374,19.023231)
\psset{linestyle=solid}
\setlinejoinmode{0}
\setlinecaps{0}
\newrgbcolor{dialinecolor}{0.000000 0.000000 0.000000}%
\psset{linecolor=dialinecolor}
\pspolygon*(75.818229,19.090842)(75.281348,19.246597)(75.449374,19.023231)(75.371497,18.754791)
\newrgbcolor{dialinecolor}{0.000000 0.000000 0.000000}%
\psset{linecolor=dialinecolor}
\pspolygon(75.818229,19.090842)(75.281348,19.246597)(75.449374,19.023231)(75.371497,18.754791)
\psset{linewidth=0.100000cm}
\psset{linestyle=solid}
\psset{linestyle=solid}
\setlinecaps{0}
\newrgbcolor{dialinecolor}{0.000000 0.000000 0.000000}%
\psset{linecolor=dialinecolor}
\psline(69.847100,28.886800)(75.463364,27.139885)
\psset{linestyle=solid}
\setlinejoinmode{0}
\setlinecaps{0}
\newrgbcolor{dialinecolor}{0.000000 0.000000 0.000000}%
\psset{linecolor=dialinecolor}
\pspolygon*(75.821442,27.028507)(75.418257,27.415730)(75.463364,27.139885)(75.269752,26.938293)
\newrgbcolor{dialinecolor}{0.000000 0.000000 0.000000}%
\psset{linecolor=dialinecolor}
\pspolygon(75.821442,27.028507)(75.418257,27.415730)(75.463364,27.139885)(75.269752,26.938293)
\psset{linewidth=0.100000cm}
\psset{linestyle=solid}
\psset{linestyle=solid}
\setlinecaps{0}
\newrgbcolor{dialinecolor}{0.000000 0.000000 0.000000}%
\psset{linecolor=dialinecolor}
\psline(76.847439,23.053141)(76.847439,23.053141)
\setfont{Helvetica}{0.800000}
\newrgbcolor{dialinecolor}{0.000000 0.000000 0.000000}%
\psset{linecolor=dialinecolor}
\rput[l](74.364000,13.278300){\psscalebox{3 -3}{CONCATENATE}}
\setfont{Helvetica}{0.800000}
\newrgbcolor{dialinecolor}{0.000000 0.000000 0.000000}%
\psset{linecolor=dialinecolor}
\rput[l](14.707000,40.344800){\psscalebox{4 -4}{\textbf{CBOW}}}
\setfont{Helvetica}{0.800000}
\newrgbcolor{dialinecolor}{0.000000 0.000000 0.000000}%
\psset{linecolor=dialinecolor}
\rput[l](42.354900,40.344800){\psscalebox{4 -4}{\textbf{Skip-gram}}}
\setfont{Helvetica}{0.800000}
\newrgbcolor{dialinecolor}{0.000000 0.000000 0.000000}%
\psset{linecolor=dialinecolor}
\rput[l](74.104000,40.203400){\psscalebox{4 -4}{\textbf{C-CONCAT}}}
}\endpspicture